\documentclass{article}
\ifdefined\pdfsuppressptexinfo\pdfsuppressptexinfo=-1\relax\fi

\usepackage{preprint}

\usepackage{amsmath,amssymb,mathtools,amsthm}

\usepackage{booktabs,array,multirow}

\usepackage{float}
\usepackage{placeins}
\usepackage{graphicx}

\usepackage[pdfusetitle]{hyperref}
\usepackage{url}

\usepackage{microtype}
\usepackage{xcolor}
\usepackage{colortbl}

\definecolor{CitationLink}{HTML}{2E5480}
\definecolor{InternalLink}{HTML}{8B4A62}
\definecolor{WebsiteLink}{HTML}{28776F}
\hypersetup{
  pdfauthor={Xinlong Guo, Jie Zhang, Lingyou Zhou},
  colorlinks=true,
  citecolor=CitationLink,
  linkcolor=InternalLink,
  urlcolor=WebsiteLink,
  filecolor=WebsiteLink,
  linktoc=section,
  bookmarksdepth=2
}

\newcommand{\TV}{\operatorname{TV}}
\newcommand{\1}{\mathbf{1}}
\newcommand{\Aact}{\mathcal A}
\newcommand{\GammaC}{\Gamma}

\newcounter{maftcalgorithm}

\newtheorem{theorem}{Theorem}[section]
\newtheorem{proposition}[theorem]{Proposition}
\newtheorem{lemma}[theorem]{Lemma}
\newtheorem{corollary}[theorem]{Corollary}

\theoremstyle{definition}
\newtheorem{definition}[theorem]{Definition}

\title{The Activity Geometry of Discrete Generative Sampling}
\author{%
  \makebox[\dimexpr\textwidth-2\tabcolsep\relax][c]{%
    Xinlong Guo\textsuperscript{1}\qquad
    Jie Zhang\textsuperscript{1,2}\qquad
    Lingyou Zhou\textsuperscript{\textdagger,1}}\\[2pt]
  \makebox[\dimexpr\textwidth-2\tabcolsep\relax][c]{%
    \small\normalfont\textsuperscript{1}Harbin Institute of Technology, Shenzhen\quad
    \textsuperscript{2}R\&D Department, Ranplan Wireless Network Design Ltd.}\\[2pt]
  \makebox[\dimexpr\textwidth-2\tabcolsep\relax][c]{%
    \small\normalfont\ttfamily\hypersetup{urlcolor=black}%
    \href{mailto:guoxinlong@stu.hit.edu.cn}{guoxinlong@stu.hit.edu.cn},
    \{\href{mailto:jiezhang@hit.edu.cn}{jiezhang}, \href{mailto:zhoulingyou@hit.edu.cn}{zhoulingyou}\}@hit.edu.cn}\\
  \makebox[\dimexpr\textwidth-2\tabcolsep\relax][c]{%
    \small\normalfont\textsuperscript{\textdagger}Corresponding author.}}

\begin{document}
\maketitle

\phantomsection
\addcontentsline{toc}{section}{Abstract}
\begin{abstract}

Iterative refinement is a key capability of diffusion-based generation
and has motivated growing efforts to bring similar behavior to discrete
diffusion and flow models.
A growing line of work enables such refinement by allowing previously
generated tokens to be revised during sampling, thereby shaping the
coupling between successive sampling states.
Despite its growing role in sampler design, this coupling freedom
remains insufficiently characterized.
We introduce \textbf{Activity Geometry} to study these finite-step couplings through
expected transition cost, which we call activity.
We show that activity decomposes into the minimum required by the induced
marginal transport and an excess term arising from the coupling.
Using this minimum as a reference, we propose
Minimal-Activity Finite-Time Coupling (MAFTC) to realize
prescribed marginals with minimum activity without
retraining or additional model evaluations.
In the continuous-time limit, the minimum activity per unit time
converges to the minimum expected jump rate compatible with the
prescribed path.
For metric-induced Gibbs paths, we further show that the activity of
the kinetic-optimal flux admits a geometric decomposition.
We find that total activity alone can obscure coupling excess: a sampler may
approach the minimum activity required by the prescribed marginal transport
while still retaining substantial coupling excess.
In language generation, as the number of model evaluations increases, refinement-based samplers accumulate
additional trajectory activity with limited diversity gains,
whereas MAFTC reduces trajectory activity while maintaining
comparable generation quality and higher lexical diversity.

\end{abstract}

\section{Introduction}

Discrete diffusion \citep{austin2021structured,lou2024sedd} and flow
\citep{gat2024dfm,shaul2025gdp} models provide frameworks for iterative
refinement, with samplers \citep{campbell2022continuous} performing
successive token updates.
At each step, the sampler uses model predictions to specify the
probabilities of retaining current token values or replacing them
with new ones \citep{gat2024dfm,shaul2025gdp}.
In particular, sampling rules can be designed to allow previously
generated tokens to be revised, providing a mechanism for iterative
refinement.
Currently, efforts to enable iterative refinement span sampling methods for
masked and uniform-state diffusion, as well as probability-path design.
In masked diffusion \citep{shi2024simplified}, the ancestral
sampler used by MDLM \citep{sahoo2024mdlm} keeps tokens fixed
once unmasked;
ReMDM \citep{wang2025remasking} then enables these tokens to be remasked
and resampled, narrowing the sample quality gap with autoregressive
models in text generation.
Meanwhile, uniform-state diffusion
\citep{hoogeboom2021argmax,austin2021structured,sahoo2025duality}
allows generated tokens to transition directly to other token values.
$\Psi$-samplers \citep{deschenaux2026psi} further extend predictor--corrector
sampling to arbitrary noise priors and, in uniform-state models,
continue to improve sample quality as the number of sampling steps
increases.
Beyond that, probability-path design provides a complementary route:
FUDOKI \citep{wang2025fudoki} uses metric-induced paths and
kinetic-optimal velocities \citep{shaul2025gdp} for iterative
refinement in a unified model for visual understanding and image
generation.

Given the input distribution, each sampling rule induces a
\emph{coupling}, the joint distribution of successive sampling states.
These revision mechanisms provide additional coupling choices,
creating opportunities for correction while also permitting repeated
token changes.
\citet{yuan2026stochasticity} show that stochastic transitions can
preserve the prescribed time marginals while contracting sampling
errors.
At the same time, \citet{xiang2026cover} identify flip-flop oscillations
in which tokens are remasked and later restored to their previous
values, and show that these oscillatory revisions can be reduced
while preserving generation quality.
However, how coupling choices shape token movement during sampling
has not been studied systematically.

To this end, we introduce \textbf{Activity Geometry}, a framework for
analyzing finite-step couplings through expected transition cost,
which we call \emph{activity}.
Our main contributions are as follows:

\noindent\textbf{(i) Finite-time Activity Geometry.}
We formulate sampler design as the selection of a coupling between
prescribed conditional marginals. Activity Geometry provides a
minimum-activity reference for this selection, leading to the
closed-form \textbf{Minimal-Activity Finite-Time Coupling (MAFTC)} construction.
We decompose a sampler's activity into the minimum required by its
induced marginal transport and the excess introduced by its coupling,
and quantify marginal mismatch between the induced and prescribed
output marginals in total variation.
MAFTC exactly transports prescribed categorical marginals with minimum
activity under the discrete mismatch cost. We extend this optimality to
transitions between factorized conditional marginals under a weighted
Hamming cost and obtain a sampler for pretrained discrete generative
models without retraining or additional model evaluations.

\noindent\textbf{(ii) Excess activity in kinetic-optimal dynamics.}
We extend the finite-time activity decomposition to instantaneous jump
rates, separating excess activity from discretization error. On the
metric-induced Gibbs paths used by FUDOKI, kinetic-optimal dynamics can
have excess activity even though every jump approaches the endpoint.
We identify exactly which transitions account for this excess.

\noindent\textbf{(iii) Trajectory activity and generation quality.}
We find in pretrained image and language models that comparable
generation quality can be achieved with substantially fewer token
changes.
In our text experiments, MAFTC's trajectory activity decreases as
the number of sampling steps increases, whereas the two baseline
samplers, Duo+DCD and $\Psi$, accumulate more token changes.
At similar generative perplexity (Gen-PPL), MAFTC also achieves
higher corpus-level diversity than both baselines.

\section{Background: Discrete Flow Matching}

Let $\mathcal T=[V]\triangleq\{1,\ldots,V\}$ be our vocabulary. The state space
is given by $\mathcal X=\mathcal T^D$ where $D\in \mathbb{N}$ is sequence length.
We consider discrete generative modeling over sequences
$x=(x^1,\ldots,x^D)\in\mathcal X$, where
$x^i\in\mathcal T$ denotes the token at position $i$.

\paragraph{Probability paths.}
We denote the source and data probability mass functions (PMFs) on
$\mathcal X$ by $p_0$ and $p_1$, respectively, and consider a factorized
source distribution
$p_0(x)=\prod_{i=1}^D p_0^i(x^i)$.
Following discrete flow matching \citep{gat2024dfm,shaul2025gdp}, we
consider a \emph{probability path} $(p_t)_{t\in[0,1]}$ interpolating
between $p_0$ and $p_1$.
We construct this path by introducing a data
endpoint $A\sim p_1$ and marginalizing \emph{conditional probability paths}
over its realizations $a\in\mathcal X$, with
\begin{equation}
p_t(x)
=
\sum_{a\in\mathcal X} p_t(x\mid a)p_1(a),
\qquad
p_t(x\mid a)
=
\prod_{i=1}^D p_t^i(x^i\mid a^i).
\label{eq:conditional-marginal-path}
\end{equation}
We refer to $(p_t)_{t\in[0,1]}$ as the \emph{marginal probability path}.
For each coordinate $i$, the conditional probability path
$p_t^i(\cdot\mid a^i)$ interpolates between the source marginal $p_0^i$
and the point mass at $a^i$, satisfying the boundary conditions
$p_0^i(\cdot\mid a^i)=p_0^i$ and
$p_1^i(\cdot\mid a^i)=\delta_{a^i}$, where
$\delta_b(z)=\1\{z=b\}$ denotes the point mass at $b$.

\paragraph{Probability velocities and fluxes.}
We model the generative dynamics by a time-inhomogeneous continuous-time Markov chain
(CTMC) $(X_t)_{t\in[0,1]}$ whose time marginals follow the prescribed probability path,
\begin{equation}
X_t\sim p_t, \qquad t\in[0,1].
\end{equation}
Its infinitesimal dynamics are governed by a
\emph{probability velocity} $u_t$, also known as a rate matrix
\citep{gat2024dfm,shaul2025gdp}.
Using the destination-first convention, $u_t(y,x)$ denotes the
instantaneous transition rate from state $x$ to state $y$, so that for small $h>0$,
\begin{equation}
\Pr(X_{t+h}=y\mid X_t=x)
=
\delta_x(y)+h\,u_t(y,x)+o(h),
\qquad h\downarrow0,
\label{eq:ctmc-short-time-transition}
\end{equation}
where $o(h)/h\to0$,
$u_t(y,x)\ge0$ for $y\ne x$, and
$\sum_y u_t(y,x)=0$.
We say that $u_t$ generates the probability path $(p_t)_{t\in[0,1]}$ when
the associated CTMC has time marginal $p_t$ at every $t$.

We restrict the CTMC to transitions that change at most one token
at a time. For each coordinate $i$, we denote by
$u_t^i(y^i,x)$ the transition rate from $x$ to the state obtained by
replacing $x^i$ with $y^i\ne x^i$ while leaving the remaining coordinates
unchanged. We set the diagonal entry to
$u_t^i(x^i,x):=-\sum_{z\in\mathcal T\setminus\{x^i\}}u_t^i(z,x)$.
Combining these coordinatewise rates gives the probability
velocity on $\mathcal X$:
\begin{equation}
u_t(y,x)
=
\sum_{i=1}^D
u_t^i(y^i,x)
\prod_{j\ne i}\delta_{x^j}(y^j).
\label{eq:coordinatewise-velocity-decomposition}
\end{equation}
Thus, for $x\ne y$, the rate $u_t(y,x)$ is zero when $x$ and $y$
differ in more than one coordinate, and equals
$u_t^i(y^i,x)$ when they differ only at coordinate $i$.

We define the \emph{probability flux}
$j_t(y,x):=p_t(x)u_t(y,x)$, whose off-diagonal entries quantify the probability mass moving
from state $x$ to state $y$ per unit time.
The marginal evolution is governed by the continuity equation,
also known as the Kolmogorov forward equation,
\begin{equation}
\dot p_t(y)
=
\sum_{x\ne y} j_t(y,x)
-
\sum_{x\ne y} j_t(x,y),
\qquad
\dot p_t(y):=\frac{d}{dt}p_t(y).
\label{eq:continuity-equation}
\end{equation}
For a fixed probability path, this equation determines only the net
flux at each state, not the individual pairwise fluxes $j_t(y,x)$.
Consequently, different probability velocities can generate the same
probability path \citep{shaul2025gdp}.

\paragraph{Posterior prediction and marginal velocities.}

Following the marginalization construction of
\citet{gat2024dfm,shaul2025gdp}, we obtain marginal velocities by
marginalizing conditional velocities with respect to the endpoint posterior.
For each coordinate $i$, we let
$u_t^i(y^i,x^i\mid a^i)$ denote a conditional velocity generating
$p_t^i(\cdot\mid a^i)$,
and the corresponding marginal velocity is given by
\begin{equation}
u_t^i(y^i,x)
=
\sum_{a^i\in\mathcal T}
u_t^i(y^i,x^i\mid a^i)\,
p_{1\mid t}^i(a^i\mid x).
\label{eq:marginal-velocity}
\end{equation}
We then combine these marginal velocities through
Eq.~\eqref{eq:coordinatewise-velocity-decomposition}
to obtain a probability velocity $u_t$ generating the marginal
probability path.
Here $p_{1\mid t}^i(\cdot\mid x)$ is the endpoint posterior for the
$i$-th token,
\begin{equation}
p_{1\mid t}^i(a^i\mid x)
:=
\Pr(A^i=a^i\mid X_t=x)
=
\sum_{a'\in\mathcal X}
\delta_{a^i}(a'^i)
\frac{p_t(x\mid a')p_1(a')}{p_t(x)},
\qquad p_t(x)>0.
\label{eq:token-endpoint-posterior}
\end{equation}
Because these endpoint posteriors depend on the unknown data
distribution $p_1$, practical discrete flow models approximate
them with a learned factorized posterior parameterized by $\theta$,
\begin{equation}
q_\theta(a\mid x,t)
=
\prod_{i=1}^D q_{\theta,i}(a^i\mid x,t),
\label{eq:factorized-posterior}
\end{equation}
where each $q_{\theta,i}(\cdot\mid x,t)$ conditions on the full current
state $x$ and approximates $p_{1\mid t}^i(\cdot\mid x)$.

\paragraph{Sampling on a time grid.}

For a PMF $\nu$ and Markov kernel $K$ on $\mathcal X$, we write
$(\nu K)(y):=\sum_x\nu(x)K(y\mid x)$ for the marginal obtained by
applying $K$ to $\nu$.
For $0\le t<s\le1$, we denote the CTMC transition kernel by
$P_{s\mid t}(y\mid x):=\Pr(X_s=y\mid X_t=x)$,
which satisfies $p_tP_{s\mid t}=p_s$.
Practical samplers operate on a grid of $N$ sampling intervals,
$0=t_0<\cdots<t_N=1$, using kernels $K_{t_n,t_{n+1}}$ constructed
from model predictions. One approach approximates a chosen CTMC
transition $P_{t_{n+1}\mid t_n}$.
Euler-type schemes
\citep{campbell2022continuous,shaul2025gdp}
freeze the transition rates at the beginning of each interval, whereas
corrected schemes
\citep{wan2026corrected,fu2026trcie}
incorporate within-interval corrections to this approximation.
Another approach selects a finite-time coupling directly between prescribed
conditional marginals, as MAFTC does below.
Applying $K_{t_n,t_{n+1}}$ to the prescribed time marginal $p_{t_n}$ gives
the \emph{induced output marginal}
\begin{equation}
\widetilde p_{t_{n+1}}
=
p_{t_n}K_{t_n,t_{n+1}}.
\label{eq:induced-output-marginal}
\end{equation}
During rollout, however, the same kernel acts on the sampler's current time
marginal, which may already differ from $p_{t_n}$.
We use the number of function evaluations (NFE) to measure the sampling
budget; Appendix~\ref{app:experiments} specifies the counting
convention under classifier-free guidance.

\section{Finite-Time Activity Geometry}\label{sec:finite-activity}

We develop Activity Geometry by relating the expected cost of a finite-time transition
to its input and output marginals. We first separate the minimum activity required
by the induced marginal transport from coupling excess and relate activity to
marginal mismatch. We then construct MAFTC and extend it to factorized conditional
probability paths.

\subsection{Transition kernels and a minimum-activity reference}
We first consider a single categorical variable with state space
$\mathcal T$, and extend the construction to sequences in
$\mathcal X=\mathcal T^D$ in Section~\ref{sec:sequence}.
Let $p$ and $q$ denote the input and prescribed output marginals, respectively.
We consider a Markov kernel $K$ on $\mathcal T$, where
$K(y\mid x)\ge0$ for all $x,y\in\mathcal T$ and
$\sum_{y\in\mathcal T}K(y\mid x)=1$ for every $x\in\mathcal T$.

\begin{definition}[Finite-time transition and activity]\label{def:finite-transition}
For $X\sim p$ and $Y\mid X\sim K(\cdot\mid X)$, we define the \emph{induced coupling} $\pi_K$ and induced output marginal $\widetilde q$ by
\[
\pi_K(x,y)=p(x)K(y\mid x),\qquad
\widetilde q(y)=(pK)(y)=\sum_x p(x)K(y\mid x).
\]
For a nonnegative transition cost $d:\mathcal T\times\mathcal T\to[0,\infty)$,
the \emph{activity} of $K$ under input
marginal $p$ is $A_d(K):=\mathbb E_{\pi_K}[d(X,Y)]$.
For the discrete mismatch cost $d(x,y)=\1\{x\ne y\}$,
we abbreviate this to $A(K)=\Pr_{\pi_K}(X\ne Y)$.
\end{definition}

Let $\GammaC(p,q)$ denote the set of couplings of $p$ and $q$.
By the classical maximal-coupling identity
\citep{griffeath1975maximal,thorisson2000coupling,jacobcouplingsmontecarlo}
the \emph{minimum activity} required to transport $p$ to $q$ is
\begin{equation}
A_{\min}(p,q):=\min_{\pi\in\GammaC(p,q)}\Pr_\pi(X\ne Y)
=\TV(p,q)=\frac12\sum_x|p(x)-q(x)|,
\label{eq:one-step-minimum}
\end{equation}
where $\TV$ denotes the total variation distance.
A maximal coupling assigns mass
$\min\{p(x),q(x)\}$ to $X=Y=x$ for each state $x$, so the remaining
probability mass corresponds to state changes. Appendix~\ref{app:proofs}
gives the derivation and characterizes the minimizing couplings.

\subsection{Induced marginal, marginal mismatch, and coupling excess}
A Markov kernel need not realize the prescribed output marginal $q$.
With $p$ and $q$ fixed, we distinguish the minimum activity needed to reach
the induced marginal $\widetilde q=pK$ from that needed to reach $q$:
\begin{equation}
A_{\rm ind}(K):=A_{\min}(p,\widetilde q),\qquad
A_{\rm pres}:=A_{\min}(p,q),\qquad
\epsilon(K):=\TV(\widetilde q,q).
\label{eq:activity-coordinates}
\end{equation}
The \emph{marginal mismatch} $\epsilon(K)$ measures the total variation
distance between the induced output marginal $\widetilde q$ and the
prescribed output marginal $q$. Since
$\pi_K\in\GammaC(p,\widetilde q)$, Eq.~\eqref{eq:one-step-minimum} gives
$A(K)=\Pr_{\pi_K}(X\ne Y)\ge A_{\min}(p,\widetilde q)=A_{\rm ind}(K)$.
We define the \emph{coupling-excess activity} as
$\chi(K):=A(K)-A_{\rm ind}(K)$, obtaining
\begin{equation}
A(K)=A_{\rm ind}(K)+\chi(K),\qquad \chi(K)\ge0.
\label{eq:defect-decomposition}
\end{equation}
The decomposition separates the minimum activity needed to reach
$\widetilde q$ from additional changes induced by the coupling;
$\epsilon(K)$ measures marginal mismatch separately. In particular,
activity below $A_{\rm pres}$ implies nonzero marginal mismatch.
We use $[z]_+:=\max\{z,0\}$ to denote the positive part of a scalar $z$.

\begin{proposition}[Marginal mismatch bound]\label{prop:defect}
For PMFs $p,q$ on $\mathcal T$ and any Markov kernel $K$ on $\mathcal T$,
\begin{equation}
A(K)+\epsilon(K)\ge A_{\rm pres},\qquad
\epsilon(K)\ge[A_{\rm pres}-A(K)]_+.
\label{eq:defect-certificate}
\end{equation}
For every $b\in[0,A_{\rm pres}]$, there is a Markov kernel $K_b$ with
$A(K_b)=b$ and $\epsilon(K_b)=A_{\rm pres}-b$.
\end{proposition}

Total activity close to $A_{\rm pres}$ can still contain substantial coupling excess when the induced marginal transition requires less activity. Appendix~\ref{app:proofs} proves the bound and its sharpness.

\subsection{Minimal-Activity Finite-Time Coupling}
We now construct a kernel that maps $p$ to $q$ with activity $A_{\rm pres}$,
so that both $\epsilon(K)$ and $\chi(K)$ vanish.
At each state $x$, we retain the common mass $c(x):=\min\{p(x),q(x)\}$
and define the \emph{loss and gain residuals} by
$r^-(x):=[p(x)-q(x)]_+$ and $r^+(x):=[q(x)-p(x)]_+$.
These are the masses to remove from and add to state $x$, respectively.
Since $p$ and $q$ both sum to one, the total mass removed equals the
total mass added:
$\tau:=\sum_x r^-(x)=\sum_x r^+(x)=\TV(p,q)$.
For $\tau>0$, we retain $c(x)$ at each state and distribute the mass
$r^-(x)$ from each loss state $x$ among gain states $y$ in proportion to
$r^+(y)$. The resulting coupling $\pi^{\rm MA}$, with MA denoting minimum
activity, is
\begin{equation}
\pi^{\rm MA}(x,x)=c(x),\qquad
\pi^{\rm MA}(x,y)=\frac{r^-(x)r^+(y)}{\tau}\quad(x\ne y).
\label{eq:ma-coupling}
\end{equation}
We call this construction \emph{Minimal-Activity Finite-Time Coupling (MAFTC)}. For $\tau>0$ and $p(x)>0$, normalizing each row yields the Markov kernel
\begin{equation}
K^{\rm MA}(y\mid x)=
\begin{cases}
\min\{p(x),q(x)\}/p(x),&y=x,\\
r^-(x)r^+(y)/(\tau p(x)),&y\ne x.
\end{cases}
\label{eq:ma-kernel}
\end{equation}
If $\tau=0$, then $p=q$, and we use the diagonal coupling
$\pi^{\rm MA}(x,y)=p(x)\delta_x(y)$ and identity kernel
$K^{\rm MA}(y\mid x)=\delta_x(y)$.
For $\tau>0$, we also set $K^{\rm MA}(\cdot\mid x)=\delta_x$ on rows
with $p(x)=0$, which do not affect the induced coupling.

\begin{proposition}[Marginal exactness and activity optimality of MAFTC]\label{prop:exact}
For PMFs $p,q$ on $\mathcal T$, the MAFTC coupling has marginals $p,q$ and mismatch probability $\TV(p,q)$. Its kernel satisfies $pK^{\rm MA}=q$ and $\epsilon(K^{\rm MA})=\chi(K^{\rm MA})=0$.
\end{proposition}

For $p(x)>0$, the kernel retains $x$ with probability $\min\{1,q(x)/p(x)\}$. When a move occurs, it draws the destination from $r^+/\tau$. This rule requires $O(V)$ work to form and sample from the residual distribution, without constructing a $V\times V$ coupling matrix. Appendix~\ref{app:proofs} proves the proposition.

\subsection{Token sequences and conditional sampling}\label{sec:sequence}
We extend the categorical construction to sequences in $\mathcal X=\mathcal T^D$ using an additive cost. For positive coordinate weights $w_1,\ldots,w_D$, define the \emph{weighted Hamming distance} $d_{H,w}(x,y):=\sum_{i=1}^D w_i\1\{x^i\ne y^i\}$.

\begin{proposition}[Minimum weighted Hamming activity]\label{prop:tensor}
Let $p=\bigotimes_{i=1}^D p^i$ and $q=\bigotimes_{i=1}^D q^i$ be PMFs on $\mathcal X$, where $p^i,q^i$ are their coordinate marginals. Every coupling $\Pi\in\GammaC(p,q)$ satisfies
\begin{equation}
\mathbb E_\Pi[d_{H,w}(X,Y)]\ge\sum_{i=1}^D w_i\TV(p^i,q^i).
\label{eq:tensor}
\end{equation}
The product of coordinatewise MAFTC couplings attains equality and is therefore optimal over all sequence couplings, including those with dependence across coordinates.
\end{proposition}

We apply this result to the factorized conditional probability path in Eq.~\eqref{eq:conditional-marginal-path}. Fix a data endpoint $a\in\mathcal X$ and sampling times $0\le t<s\le1$. For each coordinate $i$, construct $K_i^{\rm MA}$ from the time marginals $p_t^i(\cdot\mid a^i)$ and $p_s^i(\cdot\mid a^i)$. Their product, $K^a_{t,s}(y\mid x):=\prod_{i=1}^D K_i^{\rm MA}(y^i\mid x^i)$, then satisfies
\begin{equation}
p_t(\cdot\mid a)K^a_{t,s}=p_s(\cdot\mid a),
\label{eq:ma-exact}
\end{equation}
and minimizes expected weighted Hamming distance between these conditional time marginals. Appendix~\ref{app:proofs} gives the proof and a coordinatewise bound on activity and marginal mismatch without requiring factorized marginals.

The minimum-activity guarantee above applies to each fixed-endpoint transition.
To recover the marginal path, we average $K^a_{t,s}$ with the exact joint
\emph{endpoint posterior} $p_{1\mid t}(a\mid x):=\Pr(A=a\mid X_t=x)$.
For $p_t(x)>0$, Bayes' rule gives
$p_t(x)p_{1\mid t}(a\mid x)=p_1(a)p_t(x\mid a)$; hence the kernel
$K^\star_{t,s}(y\mid x):=\sum_a p_{1\mid t}(a\mid x)K^a_{t,s}(y\mid x)$
satisfies $p_tK^\star_{t,s}=p_s$.
Algorithm~\ref{alg:maftc} instead samples endpoint tokens independently from
the learned factorized posterior $q_\theta$ and applies the conditional kernels
in parallel, using one model evaluation per interval. We denote the sampler
state at $t_n$ by $\widehat X_n$ and an optional conditioning input by $\xi$.
Appendix~\ref{app:posterior} gives the resulting kernel and proves exactness
of the joint-posterior construction; Appendix~\ref{app:implementation}
describes the implementation.

\begin{figure}[!htbp]
\refstepcounter{maftcalgorithm}\label{alg:maftc}
\hrule
\vspace{3pt}
\textbf{Algorithm \themaftcalgorithm: MAFTC sampling}
\vspace{3pt}
\hrule
\vspace{3pt}
\normalsize

\begin{tabular}{@{}r@{\hspace{0.7em}}p{0.90\linewidth}@{}}
& \textbf{Input:} conditioning input $\xi$, source $p_0$, grid $0=t_0<\cdots<t_N=1$, learned token posteriors $q_{\theta,i}$, conditional probability paths $p_t^i(\cdot\mid a^i)$ \\
1 & Sample $\widehat X_0\sim p_0$. \\
2 & \textbf{for} $n=0,\ldots,N-1$ \textbf{do} \\
3 & \quad Obtain $q_{\theta,i}(\cdot\mid\widehat X_n,t_n,\xi)$ for all $i$ with one model evaluation. \\
4 & \quad Independently sample $a^i\sim q_{\theta,i}(\cdot\mid\widehat X_n,t_n,\xi)$ for all $i$; set $a=(a^1,\ldots,a^D)$. \\
5 & \quad \textbf{for} $i=1,\ldots,D$ \textbf{in parallel do} \\
6 & \qquad Set $p\gets p_{t_n}^i(\cdot\mid a^i)$, $q\gets p_{t_{n+1}}^i(\cdot\mid a^i)$, and $x\gets\widehat X_n^i$. \\
7 & \qquad Set $\tau\gets\TV(p,q)$; \textbf{if} $\tau=0$ or $p(x)=0$, set $\widehat X_{n+1}^i\gets x$ and \textbf{continue}. \\
8 & \qquad Set $\kappa\gets\min\{1,q(x)/p(x)\}$ and draw $U\sim\operatorname{Unif}[0,1]$. \\
9 & \qquad \textbf{if} $U<\kappa$, set $\widehat X_{n+1}^i\gets x$; \textbf{otherwise}, sample $\widehat X_{n+1}^i\sim[q-p]_+/\tau$. \\
10 & \quad \textbf{end for} \\
11 & \textbf{end for} \\
12 & \textbf{return} $\widehat X_N$. \\
\end{tabular}

\vspace{2pt}
\hrule
\end{figure}

\section{Activity Geometry of Kinetic-Optimal Flows}\label{sec:kinetic-activity}
We use the minimum-activity reference from Section~\ref{sec:finite-activity}
to examine whether dynamics that follow a prescribed probability path and
approach the endpoint at every jump also minimize activity.
To this end, we extend the finite-time decomposition to continuous-time
Markov dynamics, separating the minimum expected jump rate compatible
with the path from excess jump activity.
We then characterize this excess exactly for kinetic-optimal dynamics
on metric-induced Gibbs paths.

\subsection{From finite-time excess to excess jump activity}\label{sec:ct-excess}
For a CTMC, the expected jump rate is often called
\emph{dynamical activity} or \emph{traffic}
\citep{shiraishi2018speed,maes2020frenesy,vu2023thermodynamic}.
We consider a differentiable probability path $(p_t)_{t\in[0,1]}$ on
$\mathcal T$ and fix an interior time $t$.
Following the flux formulation of \citet{shaul2025gdp},
we call an off-diagonal probability flux $j_t(y,x)$ \emph{admissible}
if it is nonnegative,
vanishes whenever $p_t(x)=0$, and satisfies the continuity equation.
Its \emph{instantaneous activity} $\Aact_t(j)$ is the total probability flux
per unit time, so the constraint and objective are
\begin{equation}
\dot p_t(y)=\sum_{x\ne y}j_t(y,x)-\sum_{x\ne y}j_t(x,y),\qquad
\Aact_t(j):=\sum_{x\ne y}j_t(y,x).
\label{eq:continuity}
\end{equation}
We denote the rates of net probability gain and loss at state $x$
by $g_t(x):=[\dot p_t(x)]_+$ and $\ell_t(x):=[-\dot p_t(x)]_+$.
Because $\sum_x\dot p_t(x)=0$, their total rates coincide.
The common rate $M_t:=\sum_xg_t(x)=\sum_x\ell_t(x)=\tfrac12\|\dot p_t\|_1$
measures the probability mass that must be redistributed per unit time,
where $\|\dot p_t\|_1=\sum_x|\dot p_t(x)|$.

\begin{proposition}[Instantaneous minimum activity]\label{prop:ct}
Every admissible flux for $\dot p_t$ satisfies
\begin{equation}
\Aact_t(j)\ge M_t=\tfrac12\|\dot p_t\|_1.
\label{eq:ct-minimum}
\end{equation}
For $M_t>0$, equality is attained by
\begin{equation}
j_t^{\rm MA}(y,x)=\ell_t(x)g_t(y)/M_t,\qquad x\ne y.
\label{eq:ct-independent}
\end{equation}
Equality holds exactly when all nonzero flux moves from states with $\dot p_t<0$ to states with $\dot p_t>0$, with statewise outflow $\ell_t$ and inflow $g_t$. If $M_t=0$, the zero flux is the unique minimizer.
\end{proposition}

For a CTMC with marginal path $p_t$ and flux $j_t$, let
$P_h:=P_{t+h\mid t}$ be its exact transition kernel. Then
$p_tP_h=p_{t+h}$ and $\epsilon(P_h)=0$. Under the short-time expansion
in Eq.~\eqref{eq:ctmc-short-time-transition},
\begin{equation}
\frac{A(P_h)}{h}\longrightarrow\Aact_t(j),\qquad
\frac{\TV(p_t,p_{t+h})}{h}\longrightarrow M_t.
\label{eq:finite-infinitesimal}
\end{equation}
Thus Eq.~\eqref{eq:defect-decomposition} gives the \emph{excess jump activity}
\begin{equation}
\chi_{\rm ct}(j):=\Aact_t(j)-M_t
=\lim_{h\downarrow0}\frac{\chi(P_h)}{h}\ge0.
\label{eq:ct-excess-limit}
\end{equation}
The MAFTC coupling $\pi^{\rm MA}_{t,t+h}$ between $p_t$ and
$p_{t+h}$ satisfies $\pi^{\rm MA}_{t,t+h}(x,y)/h\to
j_t^{\rm MA}(y,x)$ for $x\ne y$ and $M_t>0$.
Appendix~\ref{app:ct-proof} proves these statements.

\subsection{Excess activity in kinetic-optimal Gibbs flows}\label{sec:gibbs-excess}
We specialize Eq.~\eqref{eq:conditional-marginal-path} to the metric-induced
Gibbs paths of \citet{shaul2025gdp}, which are also used in FUDOKI
\citep{wang2025fudoki}. Suppressing the coordinate index, we fix an
endpoint token $a\in\mathcal T$ and a finite nonnegative endpoint-distance
score $\rho(x,a)$, with $\rho(a,a)=0<\rho(x,a)$ for $x\ne a$.
Using inverse temperature $\beta\ge0$ as the path parameter, we write
\begin{equation}
p_\beta(x\mid a)=\frac{e^{-\beta \rho(x,a)}}{Z_a(\beta)},\qquad
Z_a(\beta):=\sum_{z\in\mathcal T}e^{-\beta \rho(z,a)}.
\label{eq:main-gibbs-path}
\end{equation}

The path starts at the uniform source and converges to $\delta_a$.
We write $p_\beta=p_\beta(\cdot\mid a)$ and let
$\mu(\beta):=\sum_x p_\beta(x)\rho(x,a)$ denote the average
endpoint-distance score under $p_\beta$.
For sampling times $t<s$ with inverse temperatures
$\beta_t<\beta_s$, comparing the two Gibbs marginals yields the threshold
\begin{equation}
\rho^*_{t,s}=\frac{1}{\beta_s-\beta_t}
\int_{\beta_t}^{\beta_s}\mu(b)\,db,
\label{eq:main-gibbs-threshold}
\end{equation}
which separates the supports of the finite-step loss and gain residuals
in Section~\ref{sec:finite-activity}: a state $x$ loses probability mass
when $\rho(x,a)>\rho^*_{t,s}$ and gains probability mass when
$\rho(x,a)<\rho^*_{t,s}$ (Appendix~\ref{app:kinetic}).
As $\beta_s\downarrow\beta_t$, the threshold converges to $\mu(\beta_t)$.
At fixed $\beta$, we abbreviate $\mu(\beta)$ to $\mu$.
The identity $\partial_\beta p_\beta(x)=(\mu-\rho(x,a))p_\beta(x)$
gives the corresponding instantaneous loss and gain sets;
states with $\rho(x,a)=\mu$ have zero probability derivative.

The kinetic-optimal (KO) probability velocity,
which minimizes the symmetric kinetic-energy
objective of \citet{shaul2025gdp}, induces the flux
\begin{equation}
j_\beta^{\rm KO}(y,x)
=p_\beta(x)p_\beta(y)[\rho(x,a)-\rho(y,a)]_+,\qquad x\ne y,
\label{eq:main-kinetic-flux}
\end{equation}
where the positive-part factor restricts jumps to states with smaller
endpoint-distance scores. We compare the resulting activity
$\Aact_{\rm KO}$ with the minimum $\Aact_{\rm MA}$ compatible
with the same Gibbs path.
To identify which transitions contribute to the excess jump activity
$\chi_{\rm ct}$, we use the threshold $\mu$ to partition the states into
$H:=\{x:\rho(x,a)>\mu\}$ and
$L:=\{x:\rho(x,a)\le\mu\}$, including neutral states in $L$.

\begin{theorem}[Localization of excess activity in kinetic-optimal dynamics]\label{thm:kinetic}
For the Gibbs path in Eq.~\eqref{eq:main-gibbs-path}, the total kinetic-optimal flux from $H$ to $L$ equals the minimum activity:
\begin{equation}
\sum_{x\in H,\,y\in L}j_\beta^{\rm KO}(y,x)=\Aact_{\rm MA}.
\label{eq:kinetic-cross-main}
\end{equation}
The excess jump activity equals the total flux within $H$ and $L$:
\begin{equation}
\chi_{\rm ct}(j_\beta^{\rm KO})=\Aact_{\rm KO}-\Aact_{\rm MA}
=\sum_{\substack{x,y\in H\\x\ne y}}j_\beta^{\rm KO}(y,x)
+\sum_{\substack{x,y\in L\\x\ne y}}j_\beta^{\rm KO}(y,x).
\label{eq:kinetic-excess}
\end{equation}
\end{theorem}

The theorem shows that excess jump activity can arise from probability
transport through intermediate states, even when every jump decreases
the endpoint-distance score. Eliminating the within-group transitions
while preserving the Gibbs path requires redistributing the cross-group
flux so that each state's probability derivative is unchanged.

Reparameterizing the path by a differentiable nondecreasing schedule
$\beta_t$ scales all fluxes and activity rates by $\dot\beta_t$,
preserving the decomposition wherever $\beta_t<\infty$
(Appendix~\ref{app:kinetic}).

\section{Experiments}\label{sec:experiments}
We first evaluate the activity decomposition and the localization of
excess activity in kinetic-optimal dynamics. We then compare MAFTC with baseline samplers
in terms of sampling accuracy, generation quality, diversity, and inference cost.
Appendix~\ref{app:experiments} provides complete protocols and results.
\paragraph{Setup.}
On frozen FUDOKI \citep{wang2025fudoki}, we compare MAFTC with
Euler and time-corrected (TC) sampling \citep{wan2026corrected},
both obtained by discretizing continuous-time jump rates.
On a Duo checkpoint obtained by Discrete Consistency Distillation
(DCD) \citep{sahoo2025duality}, we compare MAFTC with ancestral
sampling (Duo+DCD) and $\Psi$ \citep{deschenaux2026psi},
both derived from finite-time conditional posteriors.
Within each model, methods share the checkpoint, path, grid,
truncation, and decoding settings at matched NFE.
We evaluate image generation using GenEval, reporting the average accuracy
over its six tasks. Each image configuration contains four samples for each
of the 553 GenEval prompts \citep{ghosh2023geneval}.
For text, we report generative perplexity (Gen-PPL) under GPT-2-large
and pooled Dist-4, the number of distinct token four-grams across all
generated sequences divided by the total number of four-gram occurrences
in those sequences. Each text configuration contains 2,048 sequences
of 1,024 tokens.

\paragraph{Trajectory measurements.}
To describe realized token changes over a rollout, we use the unweighted
Hamming distance $d_H(x,y):=\sum_{i=1}^D\1\{x^i\ne y^i\}$. We define
the \emph{trajectory activity} $L_{\rm act}$, \emph{net displacement}
$D_{\rm net}$, and \emph{trajectory excess} $R_{\rm traj}$ by
\begin{equation}
L_{\rm act}=\sum_{n=0}^{N-1}d_H(\widehat X_n,\widehat X_{n+1}),\qquad
D_{\rm net}=d_H(\widehat X_0,\widehat X_N),\qquad
R_{\rm traj}=L_{\rm act}-D_{\rm net}\ge0.
\label{eq:trajectory-excess}
\end{equation}
The triangle inequality makes $R_{\rm traj}$ nonnegative. It measures Hamming path
length beyond the initial-to-final distance and can be positive even when each
transition minimizes activity, because a token can change multiple times over the
trajectory. These quantities describe individual sampled trajectories,
whereas $A(K)$ and $\chi(K)$ are expectations under a specified input marginal and
kernel. Averages of these trajectory quantities describe rollouts under
the sampler's current time marginals, which may differ from the prescribed
time marginals. For each generated sample, $L_{\rm act}$ counts token changes
over the entire sampling trajectory. We report the average of these counts
over generated samples as NTC. NFE counts model evaluations.

\subsection{Activity decomposition and localization of excess activity}
On FUDOKI's conditional paths at 32 NFE, Euler's total activity nearly matches the prescribed
minimum, yet coupling excess accounts for 24.7\% of the total because its
induced minimum is substantially lower (Figure~\ref{fig:real-models}(a)).
For the corresponding continuous-time kinetic-optimal flow,
every jump decreases endpoint distance, yet transitions within
$H$ or within $L$ account for 24.3\% of its integrated activity
(Theorem~\ref{thm:kinetic}; Appendix~\ref{app:integrated-activity}).

\subsection{Terminal accuracy and generation performance}
\paragraph{Exact-posterior sampling.}
We propagate the complete distribution on $\{0,1,2,3\}^3$ using exact
token posteriors and parallel updates, measuring the TV distance between
the induced and prescribed terminal distributions.
Figure~\ref{fig:real-models}(b) compares terminal error and activity on Gibbs
paths whose prescribed terminal distributions have dependent coordinates
(Appendix~\ref{app:oracle-accuracy}).
Coarse Euler steps can make fewer changes while incurring greater terminal
error. At 32--128 NFE, MAFTC achieves lower terminal TV than Euler and
TC with fewer token changes.

\paragraph{Pretrained generators.}
As shown in Table~\ref{tab:generation-quality}, MAFTC achieves comparable
generation quality with fewer token changes at matched NFE in both settings.
The appendix provides complete results
(Tables~\ref{tab:full-primary}, \ref{tab:language-quality},
and~\ref{tab:text-trajectories}) and paired differences with
95\% bootstrap confidence intervals
(Table~\ref{tab:paired-summary}).

Figure~\ref{fig:real-models}(c,d) decomposes NTC into net displacement
and trajectory excess. For text, as NFE increases from 8 to 128, MAFTC's
$R_{\rm traj}$ decreases from 142.8 to 54.2, whereas Duo+DCD's increases
from 1738.0 to 4246.7 and $\Psi$'s from 1759.5 to 4826.2.
Net displacement varies little, so the baselines' activity growth mainly
reflects repeated token changes, while MAFTC's activity stays close to net
displacement. Its image activity remains below that of Euler and TC at the
budgets in Table~\ref{tab:generation-quality}.

\newpage
\paragraph{Quality and diversity.}
Lower Gen-PPL can come at the cost of reduced diversity, so we compare
diversity at similar Gen-PPL.
Rep-4 measures the fraction of four-gram occurrences that repeat an
earlier occurrence within a sequence, averaged over generated sequences.
Figure~\ref{fig:text-quality-diversity} compares pooled Dist-4 and Rep-4
against Gen-PPL. In the observed comparisons at similar Gen-PPL,
MAFTC has higher pooled Dist-4 than both baselines, while $\Psi$ has
lower Rep-4 than MAFTC. Relative to $\Psi$, MAFTC's higher corpus
coverage reflects less cross-sequence four-gram reuse.
Sequence and pooled unigram entropy also give different rankings:
pooled entropy additionally reflects differences in token usage
across outputs (Appendix~\ref{app:duo-language}).

\begingroup
\setlength{\intextsep}{6pt}
\begin{table}[H]
\caption{\textbf{Generation quality and token changes at matched NFE.} Higher
is better for GenEval ($\uparrow$) and pooled Dist-4 ($\uparrow$);
lower is better for Gen-PPL ($\downarrow$).
NTC denotes the average cumulative number of token changes per sample,
rounded to the nearest integer.
Bold and underlining indicate the best and second-best GenEval,
Gen-PPL, and NTC values at each NFE.}
\label{tab:generation-quality}
\centering
\fontsize{8}{9.5}\selectfont
\setlength{\tabcolsep}{1.5pt}
\renewcommand{\arraystretch}{1.15}
\newcommand{\methodhead}[1]{%
  \begingroup\setbox2=\hbox{#1}%
  \hskip0pt plus-.5fill\kern2pt
  \leaders\hrule height-2.75pt depth3pt\hskip0pt plus.5fill
  \rlap{\lower3pt\hbox{\rule{\wd2}{0.25pt}}}\box2
  \leaders\hrule height-2.75pt depth3pt\hskip0pt plus.5fill
  \kern2pt\hskip0pt plus-.5fill\kern0pt\endgroup}
\begin{tabular}{@{}c rr rr >{\columncolor{gray!10}}r >{\columncolor{gray!10}}r rrr rrr >{\columncolor{gray!10}}r >{\columncolor{gray!10}}r >{\columncolor{gray!10}[\tabcolsep][0pt]}r@{}}
\toprule
 & \multicolumn{6}{c}{FUDOKI (Image)} & \multicolumn{9}{c}{Duo (Text)}\\
\cmidrule(lr){2-7}\cmidrule(lr){8-16}
 & \multicolumn{2}{c}{\methodhead{Euler}} & \multicolumn{2}{c}{\methodhead{TC}} & \multicolumn{2}{>{\columncolor{gray!10}}c}{\methodhead{MAFTC}} & \multicolumn{3}{c}{\methodhead{Duo+DCD}} & \multicolumn{3}{c}{\methodhead{$\Psi$}} & \multicolumn{3}{>{\columncolor{gray!10}}c}{\methodhead{MAFTC}}\\
\rule[-4pt]{0pt}{12pt}NFE & GenEval & NTC & GenEval & NTC & GenEval & NTC & Gen-PPL & NTC & Dist-4 & Gen-PPL & NTC & Dist-4 & Gen-PPL & NTC & Dist-4\\
8 & 0.690 & \underline{2146} & \textbf{0.722} & 2277 & \underline{0.712} & \textbf{1880} & \underline{59.65} & \underline{2761} & 0.8624 & \textbf{59.37} & 2783 & 0.8622 & 60.64 & \textbf{1114} & 0.8657\\
32 & \textbf{0.753} & \underline{2408} & \underline{0.745} & 2450 & 0.744 & \textbf{1883} & 34.07 & \underline{4044} & 0.8497 & \textbf{33.51} & 4177 & 0.8493 & \underline{33.59} & \textbf{1042} & 0.8515\\
128 & 0.727 & \underline{2618} & \textbf{0.739} & 2649 & \underline{0.737} & \textbf{2067} & 29.77 & \underline{5269} & 0.8461 & \textbf{27.77} & 5849 & 0.8390 & \underline{28.39} & \textbf{1029} & 0.8437\\
\bottomrule
\end{tabular}
\end{table}

\begin{figure}[H]
\centering
\includegraphics[width=\linewidth]{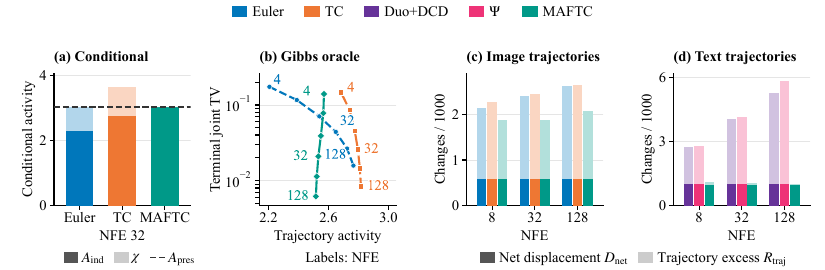}
\caption{\textbf{Activity decomposition, terminal accuracy, and sampling
trajectories.} (a) Activity decomposition on FUDOKI's conditional paths at 32 NFE.
Bars show the induced minimum $A_{\rm ind}$ and coupling excess $\chi$,
and the dashed line indicates the prescribed minimum $A_{\rm pres}$.
Values are summed over sampling intervals and averaged over endpoints.
(b) Terminal TV versus NTC on Gibbs paths at 4--128 NFE.
(c,d) Average image and text changes at 8, 32,
and 128 NFE, split into net displacement $D_{\rm net}$ and trajectory
excess $R_{\rm traj}$.}
\label{fig:real-models}
\end{figure}

\begin{figure}[H]
\centering
\includegraphics[width=\linewidth]{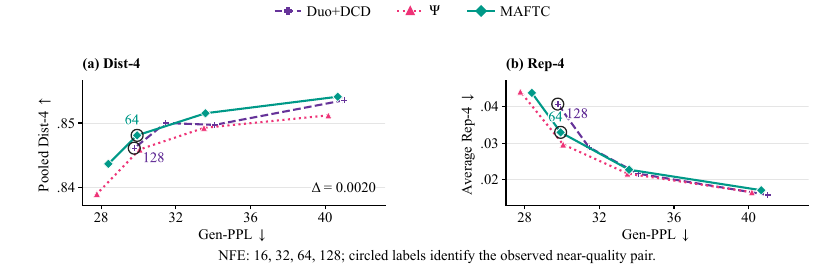}
\caption{\textbf{Corpus coverage and within-sequence repetition at similar Gen-PPL.}
Pooled Dist-4 and Rep-4 versus Gen-PPL at NFE 16, 32, 64, and 128.
Open circles identify MAFTC-64 and Duo+DCD-128 at similar Gen-PPL;
their pooled Dist-4 difference is approximately 0.0020.}
\label{fig:text-quality-diversity}
\end{figure}
\endgroup

\subsection{From transition structure to execution cost}
On the uniform text path, MAFTC uses a gate determined by the
sampling times to select positions for endpoint prediction.
Unselected positions retain their tokens, while selected positions
may retain or change them. With shared gate randomness, every
position queried by MAFTC is also queried by the ancestral and
$\Psi$ samplers (Appendix~\ref{app:execution}).
We use the same selective execution scheme for all samplers,
computing selected output rows in the final transformer block
and vocabulary head with full attention context.
We evaluate each configuration's quality and cost
(Table~\ref{tab:inference-cost}). Relative to Duo+DCD, MAFTC saves 40.87 ms
per request at 64 NFE (95\% CI $[39.66,42.07]$) and 48.38 ms at 128 NFE
($[47.79,48.94]$). Appendix~\ref{app:execution} compares dense evaluation with row selection
at the vocabulary head alone or at both the final block and vocabulary head.

\begin{table}[!htbp]
\centering
\caption{\textbf{Generation quality and inference cost per request.}
RTX 4090, seed 3407, batch size one, 1,024 tokens per sequence.
Gen-PPL is evaluated on 512 generated sequences per configuration.
Latency and TFLOPs from matrix operations are averaged over
80 timed requests; throughput is measured in final token positions per second.
Compilation and warm-up are excluded.}
\label{tab:inference-cost}
\fontsize{9}{10.8}\selectfont
\setlength{\tabcolsep}{3pt}
\renewcommand{\arraystretch}{1.04}
\begin{tabular*}{\linewidth}{@{\extracolsep{\fill}}rlrrrr@{}}
\toprule
NFE & Sampler & Gen-PPL & ms & Positions/s & TFLOPs \\
\midrule
64 & Duo+DCD & 31.406 & \underline{314.02} & \underline{3,260.9} & \underline{13.790} \\
64 & $\Psi$ & \underline{31.053} & 318.53 & 3,214.8 & 13.840 \\
64 & \textbf{MAFTC} & \textbf{30.592} & \textbf{273.16} & \textbf{3,748.8} & \textbf{13.335} \\
\addlinespace[2pt]
128 & Duo+DCD & 29.533 & \underline{554.36} & \underline{1,847.2} & \underline{26.713} \\
128 & $\Psi$ & \textbf{27.843} & 558.34 & 1,834.0 & 26.800 \\
128 & \textbf{MAFTC} & \underline{28.867} & \textbf{505.98} & \textbf{2,023.8} & \textbf{26.216} \\
\bottomrule
\end{tabular*}
\end{table}

\FloatBarrier

\section{Related Work}
\paragraph{Jump objectives and finite-time couplings.}
\citet{campbell2024discreteflows} construct jump-minimizing generators for uniform
and masking conditional paths, whereas \citet{haxholli2026minibatch} optimize
source--target training couplings under a jump-cost objective.
For prescribed paths, General Discrete Paths \citep{shaul2025gdp} minimizes
symmetric kinetic energy; we identify the transitions responsible for excess
activity in the resulting dynamics on metric-induced Gibbs paths.
At finite steps, maximal coupling appears in learned noising processes
\citep{bartosh2026forward} and
uniform-diffusion bridges \citep{song2021ddim,gourevitch2026uniform}.
MAFTC applies this principle to prescribed conditional marginals.
Our framework decomposes activity into the minimum required by induced marginal
transport and coupling excess, while measuring marginal mismatch separately.

\paragraph{Revision and error correction.}
ReMDM \citep{wang2025remasking} and $\Psi$-samplers \citep{deschenaux2026psi} enable revision through predictor--corrector sampling. \citet{yuan2026stochasticity} analyze marginal-preserving stochastic transitions that contract sampling errors, whereas COVER \citep{xiang2026cover} reduces remask--restore oscillations. We quantify changes relative to the minimum activity required by marginal transport, providing a distributional reference for interpreting revision activity.

\paragraph{Sampling accuracy.}
Corrected samplers \citep{wan2026corrected,fu2026trcie} improve finite-step approximation, while evaluations with exact token posteriors \citep{tang2026sampler} isolate sampler error from learned prediction error. We complement these studies by examining sampling accuracy and generation quality alongside activity.

\section{Conclusion}
Activity Geometry formulates sampler design through couplings of
prescribed conditional marginals. Its minimum-activity reference
yields MAFTC and reveals excess in kinetic-optimal Gibbs dynamics
that preserve the path while approaching the endpoint.
On finer grids, MAFTC achieves comparable image and text quality
with fewer token changes. Its text trajectory excess decreases
with grid refinement, while it increases for both baselines.
Selective execution translates its transition structure into
lower inference cost.

\clearpage
\subsection*{AI use statement}
\addcontentsline{toc}{section}{AI use statement}
Generative AI tools, including OpenAI ChatGPT, assisted with literature organization, brainstorming, mathematical consistency checks, language editing, and code and experiment analysis. The authors verified mathematical claims through explicit derivations, checked literature claims against the cited sources, and reviewed and tested code changes before use. All reported experiments and numerical results were produced from author-controlled code and experiment logs. The authors reviewed the complete manuscript and take responsibility for its content.

\subsection*{Reproducibility statement}
\addcontentsline{toc}{section}{Reproducibility statement}
Appendices~\ref{app:proofs} and~\ref{app:kinetic} give complete proofs of the main claims and state their finite-state assumptions. Appendix~\ref{app:implementation} describes the vectorized MAFTC update, numerical stability, and computational complexity. Appendix~\ref{app:experiments} records the frozen checkpoints, sampling configurations, image and language metrics, activity definitions, and paired-bootstrap protocol. The ancillary file \texttt{supplementary\_tables.zip} contains the numerical results referenced in the experimental appendices.

\renewcommand{\bibsection}{%
  \section*{\refname}%
  \addcontentsline{toc}{section}{\refname}%
}
\bibliographystyle{plainnat}
\bibliography{references}

@inproceedings{austin2021structured,
  title     = {Structured Denoising Diffusion Models in Discrete State-Spaces},
  author    = {Austin, Jacob and Johnson, Daniel D. and Ho, Jonathan and Tarlow, Daniel and van den Berg, Rianne},
  booktitle = {Advances in Neural Information Processing Systems},
  volume    = {34},
  year      = {2021}
}

@article{bartosh2026forward,
  title     = {Forward-Learned Discrete Diffusion: Learning How to Noise to Denoise Faster},
  author    = {Bartosh, Grigory and Pandeva, Teodora and Karmalkar, Sushrut and Zazo, Javier},
  journal   = {arXiv preprint arXiv:2605.18204},
  year      = {2026},
  url       = {https://arxiv.org/abs/2605.18204}
}

@inproceedings{campbell2022continuous,
  title     = {A Continuous Time Framework for Discrete Denoising Models},
  author    = {Campbell, Andrew and Benton, Joe and De Bortoli, Valentin and Rainforth, Thomas and Deligiannidis, George and Doucet, Arnaud},
  booktitle = {Advances in Neural Information Processing Systems},
  volume    = {35},
  year      = {2022}
}

@inproceedings{campbell2024discreteflows,
  title     = {Generative Flows on Discrete State-Spaces: Enabling Multimodal Flows with Applications to Protein Co-Design},
  author    = {Campbell, Andrew and Yim, Jason and Barzilay, Regina and Rainforth, Tom and Jaakkola, Tommi},
  booktitle = {Proceedings of the 41st International Conference on Machine Learning},
  series    = {Proceedings of Machine Learning Research},
  volume    = {235},
  pages     = {5453--5512},
  publisher = {PMLR},
  year      = {2024},
  url       = {https://proceedings.mlr.press/v235/campbell24a.html}
}

@article{deschenaux2026psi,
  title     = {The Diffusion Duality, Chapter {II}: {$\Psi$}-Samplers and Efficient Curriculum},
  author    = {Deschenaux, Justin and Gulcehre, Caglar and Sahoo, Subham Sekhar},
  journal   = {arXiv preprint arXiv:2602.21185v1},
  year      = {2026},
  url       = {https://arxiv.org/abs/2602.21185v1}
}

@inproceedings{fu2026trcie,
  title     = {A Time-Reparameterized Cumulative Intensity Extrapolation Sampler for Discrete Flow Matching},
  author    = {Fu, Feiyang and Fan, Hehe},
  booktitle = {International Conference on Machine Learning},
  year      = {2026}
}

@inproceedings{gat2024dfm,
  title     = {Discrete Flow Matching},
  author    = {Gat, Itai and Remez, Tal and Shaul, Neta and Kreuk, Felix and Chen, Ricky T. Q. and Synnaeve, Gabriel and Adi, Yossi and Lipman, Yaron},
  booktitle = {Advances in Neural Information Processing Systems},
  volume    = {37},
  year      = {2024}
}

@inproceedings{ghosh2023geneval,
  title     = {{GenEval}: An Object-Focused Framework for Evaluating Text-to-Image Alignment},
  author    = {Ghosh, Dhruba and Hajishirzi, Hannaneh and Schmidt, Ludwig},
  booktitle = {Advances in Neural Information Processing Systems Datasets and Benchmarks Track},
  volume    = {36},
  pages     = {52132--52152},
  year      = {2023},
  url       = {https://proceedings.neurips.cc/paper_files/paper/2023/hash/a3bf71c7c63f0c3bcb7ff67c67b1e7b1-Abstract-Datasets_and_Benchmarks.html}
}

@article{gourevitch2026uniform,
  title         = {Uniform Diffusion Models Revisited: Leave-One-Out Denoiser and Absorbing State Reformulation},
  author        = {Gourevitch, Samson and Janati, Yazid and Shariatian, Dario and Simsekli, Umut and Moulines, Eric and Xing, Eric P. and Durmus, Alain},
  journal       = {arXiv preprint arXiv:2605.22765},
  year          = {2026},
  eprint        = {2605.22765},
  archivePrefix = {arXiv},
  primaryClass  = {cs.LG},
  url           = {https://arxiv.org/abs/2605.22765}
}

@article{griffeath1975maximal,
  title   = {A Maximal Coupling for {Markov} Chains},
  author  = {Griffeath, David},
  journal = {Zeitschrift f{\"u}r Wahrscheinlichkeitstheorie und Verwandte Gebiete},
  volume  = {31},
  number  = {2},
  pages   = {95--106},
  year    = {1975},
  doi     = {10.1007/BF00539434}
}

@inproceedings{haxholli2026minibatch,
  title     = {Minibatch Optimal Transport and Perplexity Bound Estimation in Discrete Flow Matching},
  author    = {Haxholli, Etrit and G{\"u}rb{\"u}z, Yeti Z. and Can, O{\u g}ul and Waxman, Eli},
  booktitle = {International Conference on Machine Learning},
  year      = {2026},
  url       = {https://openreview.net/forum?id=A8rmJlSET9}
}

@article{holtzman2020degeneration,
  title     = {The Curious Case of Neural Text Degeneration},
  author    = {Holtzman, Ari and Buys, Jan and Du, Li and Forbes, Maxwell and Choi, Yejin},
  journal   = {arXiv preprint arXiv:1904.09751v2},
  year      = {2020},
  url       = {https://arxiv.org/abs/1904.09751v2}
}

@inproceedings{hoogeboom2021argmax,
  title     = {Argmax Flows and Multinomial Diffusion: Learning Categorical Distributions},
  author    = {Hoogeboom, Emiel and Nielsen, Didrik and Jaini, Priyank and Forr{\'e}, Patrick and Welling, Max},
  booktitle = {Advances in Neural Information Processing Systems},
  volume    = {34},
  pages     = {12454--12465},
  year      = {2021},
  url       = {https://proceedings.neurips.cc/paper_files/paper/2021/file/67d96d458abdef21792e6d8e590244e7-Paper.pdf}
}

@misc{jacobcouplingsmontecarlo,
  author       = {Jacob, Pierre E.},
  title        = {Couplings and {Monte Carlo}},
  year         = {2021},
  howpublished = {Lecture notes, STAT 248, Harvard University},
  url          = {https://sites.google.com/site/pierrejacob/cmclectures}
}

@inproceedings{li2016diversity,
  title     = {A Diversity-Promoting Objective Function for Neural Conversation Models},
  author    = {Li, Jiwei and Galley, Michel and Brockett, Chris and Gao, Jianfeng and Dolan, Bill},
  booktitle = {Proceedings of the 2016 Conference of the North American Chapter of the Association for Computational Linguistics: Human Language Technologies},
  publisher = {Association for Computational Linguistics},
  pages     = {110--119},
  year      = {2016},
  doi       = {10.18653/v1/N16-1014},
  url       = {https://aclanthology.org/N16-1014/}
}

@inproceedings{liu2022rethinking,
  title     = {Rethinking and Refining the Distinct Metric},
  author    = {Liu, Siyang and Sabour, Sahand and Zheng, Yinhe and Ke, Pei and Zhu, Xiaoyan and Huang, Minlie},
  booktitle = {Proceedings of the 60th Annual Meeting of the Association for Computational Linguistics (Volume 2: Short Papers)},
  publisher = {Association for Computational Linguistics},
  pages     = {762--770},
  year      = {2022},
  doi       = {10.18653/v1/2022.acl-short.86},
  url       = {https://aclanthology.org/2022.acl-short.86/}
}

@inproceedings{lou2024sedd,
  title     = {Discrete Diffusion Modeling by Estimating the Ratios of the Data Distribution},
  author    = {Lou, Aaron and Meng, Chenlin and Ermon, Stefano},
  booktitle = {Proceedings of the 41st International Conference on Machine Learning},
  series    = {Proceedings of Machine Learning Research},
  volume    = {235},
  pages     = {32819--32848},
  year      = {2024},
  url       = {https://proceedings.mlr.press/v235/lou24a.html}
}

@inproceedings{ma2025hpsv3,
  author = {Ma, Yuhang and Shui, Yunhao and Wu, Xiaoshi and Sun, Keqiang and Li, Hongsheng},
  title = {{HPSv3}: Towards Wide-Spectrum Human Preference Score},
  booktitle = {Proceedings of the IEEE/CVF International Conference on Computer Vision},
  year = {2025},
  url = {https://arxiv.org/abs/2508.03789}
}

@article{maes2020frenesy,
  title   = {Frenesy: Time-Symmetric Dynamical Activity in Nonequilibria},
  author  = {Maes, Christian},
  journal = {Physics Reports},
  volume  = {850},
  pages   = {1--33},
  year    = {2020},
  doi     = {10.1016/j.physrep.2020.01.002}
}

@techreport{radford2019language,
  title       = {Language Models are Unsupervised Multitask Learners},
  author      = {Radford, Alec and Wu, Jeffrey and Child, Rewon and Luan, David and Amodei, Dario and Sutskever, Ilya},
  institution = {OpenAI},
  year        = {2019},
  url         = {https://cdn.openai.com/better-language-models/language_models_are_unsupervised_multitask_learners.pdf}
}

@inproceedings{radford2021clip,
  title     = {Learning Transferable Visual Models from Natural Language Supervision},
  author    = {Radford, Alec and Kim, Jong Wook and Hallacy, Chris and Ramesh, Aditya and Goh, Gabriel and Agarwal, Sandhini and Sastry, Girish and Askell, Amanda and Mishkin, Pamela and Clark, Jack and Krueger, Gretchen and Sutskever, Ilya},
  booktitle = {Proceedings of the 38th International Conference on Machine Learning},
  series    = {Proceedings of Machine Learning Research},
  volume    = {139},
  pages     = {8748--8763},
  year      = {2021}
}

@inproceedings{sahoo2024mdlm,
  title     = {Simple and Effective Masked Diffusion Language Models},
  author    = {Sahoo, Subham Sekhar and Arriola, Marianne and Schiff, Yair and Gokaslan, Aaron and Marroquin, Edgar and Chiu, Justin T. and Rush, Alexander and Kuleshov, Volodymyr},
  booktitle = {Advances in Neural Information Processing Systems},
  volume    = {37},
  year      = {2024}
}

@inproceedings{sahoo2025duality,
  title     = {The Diffusion Duality},
  author    = {Sahoo, Subham Sekhar and Deschenaux, Justin and Gokaslan, Aaron and Wang, Guanghan and Chiu, Justin T. and Kuleshov, Volodymyr},
  booktitle = {Proceedings of the 42nd International Conference on Machine Learning},
  series    = {Proceedings of Machine Learning Research},
  volume    = {267},
  pages     = {52584--52619},
  year      = {2025},
  url       = {https://proceedings.mlr.press/v267/sahoo25a.html}
}

@article{shaul2025gdp,
  title     = {Flow Matching with General Discrete Paths: A Kinetic-Optimal Perspective},
  author    = {Shaul, Neta and Gat, Itai and Havasi, Marton and Severo, Daniel and Sriram, Anuroop and Holderrieth, Peter and Karrer, Brian and Lipman, Yaron and Chen, Ricky T. Q.},
  journal   = {arXiv preprint arXiv:2412.03487},
  year      = {2024},
  url       = {https://arxiv.org/abs/2412.03487}
}

@inproceedings{shi2024simplified,
  title     = {Simplified and Generalized Masked Diffusion for Discrete Data},
  author    = {Shi, Jiaxin and Han, Kehang and Wang, Zhe and Doucet, Arnaud and Titsias, Michalis K.},
  booktitle = {Advances in Neural Information Processing Systems},
  volume    = {37},
  pages     = {103131--103167},
  year      = {2024},
  doi       = {10.52202/079017-3277},
  url       = {https://proceedings.neurips.cc/paper_files/paper/2024/hash/bad233b9849f019aead5e5cc60cef70f-Abstract-Conference.html}
}

@article{shiraishi2018speed,
  title   = {Speed Limit for Classical Stochastic Processes},
  author  = {Shiraishi, Naoto and Funo, Ken and Saito, Keiji},
  journal = {Physical Review Letters},
  volume  = {121},
  number  = {7},
  pages   = {070601},
  year    = {2018},
  doi     = {10.1103/PhysRevLett.121.070601}
}

@article{song2021ddim,
  title     = {Denoising Diffusion Implicit Models},
  author    = {Song, Jiaming and Meng, Chenlin and Ermon, Stefano},
  journal   = {arXiv preprint arXiv:2010.02502v2},
  year      = {2021},
  url       = {https://arxiv.org/abs/2010.02502v2}
}

@inproceedings{tang2026sampler,
  title     = {Is Your Diffusion Sampler Actually Correct? {A} Sampler-Centric Evaluation of Discrete Diffusion Language Models},
  author    = {Tang, Luhan and Yu, Longxuan and Zhang, Shaorong and Ver Steeg, Greg},
  booktitle = {International Conference on Machine Learning},
  year      = {2026}
}

@book{thorisson2000coupling,
  title     = {Coupling, Stationarity, and Regeneration},
  author    = {Thorisson, Hermann},
  publisher = {Springer},
  year      = {2000}
}

@article{vu2023thermodynamic,
  title   = {Thermodynamic Unification of Optimal Transport: Thermodynamic Uncertainty Relation, Minimum Dissipation, and Thermodynamic Speed Limits},
  author  = {Van Vu, Tan and Saito, Keiji},
  journal = {Physical Review X},
  volume  = {13},
  number  = {1},
  pages   = {011013},
  year    = {2023},
  doi     = {10.1103/PhysRevX.13.011013}
}

@book{villani2009ot,
  title     = {Optimal Transport: Old and New},
  author    = {Villani, C{\'e}dric},
  publisher = {Springer},
  year      = {2009},
  series    = {Grundlehren der mathematischen Wissenschaften},
  volume    = {338},
  doi       = {10.1007/978-3-540-71050-9}
}

@inproceedings{wan2026corrected,
  title     = {Corrected Samplers for Discrete Flow Models},
  author    = {Wan, Zhengyan and Ouyang, Yidong and Xie, Liyan and Zha, Hongyuan and Fang, Fang and Cheng, Guang},
  booktitle = {International Conference on Machine Learning},
  year      = {2026}
}

@inproceedings{wang2025remasking,
  title     = {Remasking Discrete Diffusion Models with Inference-Time Scaling},
  author    = {Wang, Guanghan and Schiff, Yair and Sahoo, Subham Sekhar and Kuleshov, Volodymyr},
  booktitle = {Advances in Neural Information Processing Systems},
  volume    = {38},
  pages     = {163488--163545},
  year      = {2025},
  doi       = {10.52202/085713-4927}
}

@inproceedings{wang2025fudoki,
  title     = {{FUDOKI}: Discrete Flow-based Unified Understanding and Generation via Kinetic-Optimal Velocities},
  author    = {Wang, Jin and Lai, Yao and Li, Aoxue and Zhang, Shifeng and Sun, Jiacheng and Kang, Ning and Wu, Chengyue and Li, Zhenguo and Luo, Ping},
  booktitle = {Advances in Neural Information Processing Systems},
  volume    = {38},
  year      = {2025}
}

@article{wiher2022decoding,
  title   = {On Decoding Strategies for Neural Text Generators},
  author  = {Wiher, Gian and Meister, Clara and Cotterell, Ryan},
  journal = {Transactions of the Association for Computational Linguistics},
  volume  = {10},
  pages   = {997--1012},
  year    = {2022},
  doi     = {10.1162/tacl_a_00502},
  url     = {https://aclanthology.org/2022.tacl-1.58/}
}

@article{wu2023hpsv2,
  author = {Wu, Xiaoshi and Hao, Yiming and Sun, Keqiang and Chen, Yixiong and Zhu, Feng and Zhao, Rui and Li, Hongsheng},
  title = {Human Preference Score v2: A Solid Benchmark for Evaluating Human Preferences of Text-to-Image Synthesis},
  journal = {arXiv preprint arXiv:2306.09341},
  year = {2023},
  url = {https://arxiv.org/abs/2306.09341}
}

@inproceedings{xiang2026cover,
  title         = {Stop the Flip-Flop: Context-Preserving Verification for Fast Revocable Diffusion Decoding},
  author        = {Xiang, Yanzheng and Wei, Lan and Yao, Yizhen and Zhu, Qinglin and Yan, Hanqi and Jin, Chen and Teare, Philip Alexander and Zhang, Dandan and Gui, Lin and Saseendran, Amrutha and He, Yulan},
  booktitle     = {International Conference on Machine Learning},
  year          = {2026},
  eprint        = {2602.06161},
  archivePrefix = {arXiv},
  primaryClass  = {cs.CL},
  url           = {https://arxiv.org/abs/2602.06161}
}

@article{yuan2026stochasticity,
  title = {On the Error-Correcting Effects of Stochasticity in Discrete Diffusion},
  author = {Yuan, William and Jeong, Sungwon and Aghazadeh, Amirali},
  journal = {arXiv preprint arXiv:2605.26582},
  year = {2026},
  url = {https://arxiv.org/abs/2605.26582}
}

\clearpage
\begingroup
\makeatletter
  \renewcommand{\contentsname}{CONTENTS}%
  \renewcommand{\tableofcontents}{%
    \par{\normalfont\LARGE\bfseries\color{InternalLink}\contentsname\par}%
    \nobreak\vspace{8pt}%
    \@starttoc{toc}%
  }%
  \setcounter{tocdepth}{3}%
  \normalsize
  \linespread{1.15}\selectfont
  \setlength{\parskip}{0pt}%
  \renewcommand{\l@section}[2]{%
    \addvspace{0.6em}%
    \@dottedtocline{1}{0em}{1.8em}{\textbf{#1}}{\textbf{#2}}%
  }%
  \renewcommand{\l@subsection}{\@dottedtocline{2}{1.8em}{2.3em}}%
  \renewcommand{\l@subsubsection}{\@dottedtocline{3}{4.1em}{3.2em}}%
  \let\ref\getrefnumber
  \tableofcontents
  \par

\makeatother
\endgroup
\clearpage
\setlength{\parskip}{0pt}

\appendix

\section{Proofs of Main Results}\label{app:proofs}
We establish the coupling lower bound, then prove the marginal mismatch
bound, MAFTC optimality, the sequence result, and the continuous-time limit.

\subsection{Classical maximal coupling and residual couplings}\label{app:residual-pairing}
Let $p$ and $q$ be PMFs on the state space $\mathcal T$.
The coupling set has the standard marginal constraints \citep{villani2009ot},
\[
\GammaC(p,q)=\left\{\pi\ge0:
\begin{array}{ll}
\sum_{y\in\mathcal T}\pi(x,y)=p(x),&x\in\mathcal T,\\
\sum_{x\in\mathcal T}\pi(x,y)=q(y),&y\in\mathcal T
\end{array}
\right\}.
\]
We first separate the mass shared by $p$ and $q$ from the mass that must move. Let
\[
c(x):=\min\{p(x),q(x)\},\qquad
r^-(x):=[p(x)-q(x)]_+,\qquad
r^+(x):=[q(x)-p(x)]_+.
\]
The loss residual $r^-$ is the mass to remove from the input, and the gain
residual $r^+$ is the mass to add to reach the output; they satisfy
\[
p=c+r^-,\qquad q=c+r^+,\qquad q-p=r^+-r^-.
\]
For every state $x$, at least one of the two residuals is zero, giving $r^-(x)r^+(x)=0$ and
$|p(x)-q(x)|=r^-(x)+r^+(x)$.
Moreover, since $\sum_x \bigl(q(x)-p(x)\bigr)=0,$
we have
\begin{equation}
\tau:=\sum_xr^-(x)=\sum_xr^+(x)
=1-\sum_xc(x)
=\frac12\sum_x|p(x)-q(x)|=\TV(p,q),
\label{eq:tv-overlap}
\end{equation}
where the last equality is the standard expression for total variation
distance on a discrete state space
\citep{griffeath1975maximal,thorisson2000coupling}.

\begin{lemma}[Coupling lower bound and equality conditions]\label{lem:coupling}
For every $\pi\in\GammaC(p,q)$, $\Pr_\pi(X\ne Y)\ge\TV(p,q)$.
Equality holds exactly when $\pi(x,x)=\min\{p(x),q(x)\}$ for every $x$.
\end{lemma}

\begin{proof}
Since $\pi$ has marginals $p$ and $q$, we have $\pi(x,x)\le c(x)$
for every $x$. Hence, by
Eq.~\eqref{eq:tv-overlap},
\begin{equation}
\Pr_\pi(X\ne Y)-\TV(p,q)
=
\left(1-\sum_x\pi(x,x)\right)
-
\left(1-\sum_x c(x)\right)
=
\sum_x\bigl[c(x)-\pi(x,x)\bigr]
\ge0.
\label{eq:coupling-lower-bound-proof}
\end{equation}
Thus $\Pr_\pi(X\ne Y)\ge\TV(p,q)$. Since every summand in
Eq.~\eqref{eq:coupling-lower-bound-proof} is nonnegative, equality holds
if and only if $\pi(x,x)=c(x)$ for every $x$.

We next characterize the couplings attaining equality. Suppose $\tau>0$,
and let $D_c$ be the diagonal array with
$[D_c]_{xy}:=c(x)\mathbf1\{x=y\}$. For any coupling $\pi$ attaining equality, define
\begin{equation}
\gamma(x,y)
:=
\frac{\pi(x,y)-[D_c]_{xy}}{\tau}.
\label{eq:residual-coupling-definition}
\end{equation}
The equality condition gives $\gamma(x,x)=0$, while for $x\ne y$,
$\gamma(x,y)=\pi(x,y)/\tau\ge0$. Moreover, its row and column marginals are
\begin{equation}
\sum_y\gamma(x,y)=\frac{r^-(x)}{\tau},
\qquad
\sum_x\gamma(x,y)=\frac{r^+(y)}{\tau}.
\label{eq:residual-coupling-marginals}
\end{equation}
Therefore every coupling attaining equality with $\tau>0$ has the form
\begin{equation}
\pi=D_c+\tau\gamma,
\qquad
\gamma\in\GammaC(r^-/\tau,r^+/\tau).
\label{eq:minimum-activity-family}
\end{equation}

Conversely, let $\gamma\in\GammaC(r^-/\tau,r^+/\tau)$ and define
$\pi:=D_c+\tau\gamma$. Since $r^-(x)r^+(x)=0$ for every $x$, at least one
of the row or column marginals of $\gamma$ at $x$ is zero, so
$\gamma(x,x)=0$. Hence $\pi$ has marginals $p$ and $q$ and diagonal entries
$\pi(x,x)=c(x)$. By the equality condition above, it therefore attains
$\Pr_\pi(X\ne Y)=\TV(p,q)$.

Finally, if $\tau=0$, then $p=q$, and the equality condition uniquely gives
the diagonal coupling $\pi(x,y)=p(x)\mathbf1\{x=y\}$.
\end{proof}

We also use the correspondence between couplings and Markov kernels.
For any $\pi\in\GammaC(p,q)$, define
\begin{equation}
K_\pi(y\mid x)=
\begin{cases}
\pi(x,y)/p(x),&p(x)>0,\\
\delta_x(y),&p(x)=0.
\end{cases}
\label{eq:kernel-from-coupling}
\end{equation}
Each row sums to one. A zero row marginal forces $\pi(x,y)=0$ for all $y$,
so $p(x)K_\pi(y\mid x)=\pi(x,y)$ also holds when $p(x)=0$.
Summing over $x$ gives $pK_\pi=q$. Any probability distribution
can replace the identity row on these zero-probability states without affecting
the induced coupling.

\subsection{Proof of Proposition~\ref{prop:defect} and sharpness}
\begin{proof}
The induced law $\pi_K(x,y)=p(x)K(y\mid x)$ couples $p$ and $\widetilde q=pK$. Lemma~\ref{lem:coupling} gives $A(K)\ge\TV(p,\widetilde q)$, hence $\chi(K)\ge0$. The triangle inequality gives
\[
\TV(p,q)\le\TV(p,\widetilde q)+\TV(\widetilde q,q)
\le A(K)+\epsilon(K).
\]
Rearranging and using $\epsilon(K)\ge0$ proves Eq.~\eqref{eq:defect-certificate}.

For sharpness, let $\tau=\TV(p,q)>0$ and choose an activity level $b\in[0,\tau]$. Set $\lambda=b/\tau$ and $\widetilde q=(1-\lambda)p+\lambda q$. Since $\widetilde q-p=\lambda(q-p)$ and $q-\widetilde q=(1-\lambda)(q-p)$, homogeneity of the $\ell^1$ norm gives
\[
\TV(p,\widetilde q)=\lambda\TV(p,q)=b,\qquad
\TV(\widetilde q,q)=(1-\lambda)\TV(p,q)=\tau-b.
\]
By Eq.~\eqref{eq:kernel-from-coupling}, a maximal coupling of $p$ and $\widetilde q$ therefore induces a kernel with $A(K)=b$ and $\epsilon(K)=\tau-b$. If $p=q$, the identity kernel attains zero activity and marginal mismatch.
\end{proof}

\begin{corollary}[Coordinatewise bound on activity and marginal mismatch]\label{cor:defect}
Let $\Pi$ couple sequence-valued input and output states with coordinate
marginals $p^i$ and $\widetilde q^i$, respectively. Fix prescribed output
coordinate marginals $q^i$ and positive weights $w_1,\ldots,w_D$. Then
\[
\mathbb E_\Pi[d_{H,w}(X,Y)]+\sum_iw_i\TV(\widetilde q^i,q^i)
\ge \sum_iw_i\TV(p^i,q^i).
\]
\end{corollary}

\begin{proof}[Proof of Corollary~\ref{cor:defect}]
Project $\Pi$ onto coordinate $i$ and write
$A_i:=\Pr_\Pi(X^i\ne Y^i)$. This mismatch probability and the induced
output marginal $\widetilde q^i$ satisfy Proposition~\ref{prop:defect}, so
\[
A_i+\TV(\widetilde q^i,q^i)\ge\TV(p^i,q^i).
\]
Multiplication by $w_i>0$ preserves the inequality. Summing over coordinates and using
$\mathbb E_\Pi[d_{H,w}(X,Y)]=\sum_iw_iA_i$ gives the stated bound.
\end{proof}

\subsection{MAFTC construction and proof of Proposition~\ref{prop:exact}}
\begin{proof}
Use the common mass and residuals from Appendix~\ref{app:residual-pairing}. If $\tau=0$, the identity kernel realizes $p=q$ with zero activity. For $\tau>0$, choose the independent residual coupling
\[
\gamma=(r^-/\tau)\otimes(r^+/\tau),\qquad
\gamma(x,y)=\frac{r^-(x)r^+(y)}{\tau^2},
\]
where $\otimes$ denotes the product of the two probability distributions. It belongs to $\GammaC(r^-/\tau,r^+/\tau)$, so Eq.~\eqref{eq:minimum-activity-family} gives
\[
\pi^{\rm MA}=D_c+\tau\gamma.
\]
This coupling has marginals $p,q$ and activity $\tau$. Applying Eq.~\eqref{eq:kernel-from-coupling} gives the kernel in Eq.~\eqref{eq:ma-kernel}, with $pK^{\rm MA}=q$. Hence $\epsilon(K^{\rm MA})=0$ and $\chi(K^{\rm MA})=\tau-\TV(p,q)=0$.

For a loss state $x$, $p(x)>q(x)$, so its retention and move probabilities are $q(x)/p(x)$ and $r^-(x)/p(x)$, respectively. Conditional on a move,
\[
\Pr(Y=y\mid X=x,\,Y\ne X)
=\frac{K^{\rm MA}(y\mid x)}{r^-(x)/p(x)}
=\frac{r^+(y)}{\tau},\qquad y\ne x.
\]
For a gain or neutral state with $p(x)>0$, $r^-(x)=0$ and the kernel retains the state. Zero-probability input states use the identity-row convention in Eq.~\eqref{eq:kernel-from-coupling}.
\end{proof}

\subsection{Proof of Proposition~\ref{prop:tensor}}
\begin{proof}
Let $\Pi$ be any coupling of the product distributions $p(x)=\prod_i p^i(x^i)$ and $q(y)=\prod_i q^i(y^i)$, written $p=\otimes_i p^i$ and $q=\otimes_i q^i$. We do not assume that $\Pi$ factorizes. The joint law $\Pi_i$ of $(X^i,Y^i)$ belongs to $\GammaC(p^i,q^i)$ because its marginals are $p^i$ and $q^i$. Therefore the one-coordinate result gives
\[
\mathbb E_\Pi\1\{X^i\ne Y^i\}
=\Pr_{\Pi_i}(X^i\ne Y^i)
\ge\TV(p^i,q^i).
\]
Multiplying by $w_i>0$ and summing yields
\begin{equation}
\mathbb E_\Pi[d_{H,w}(X,Y)]
=\sum_iw_i\Pr_{\Pi_i}(X^i\ne Y^i)
\ge\sum_iw_i\TV(p^i,q^i).
\label{eq:tensor-lower}
\end{equation}
For achievability, choose any maximal coupling $\pi_i^\star\in\GammaC(p^i,q^i)$ for every coordinate and form the product coupling
\[
\Pi^\star((x^1,\ldots,x^D),(y^1,\ldots,y^D))
=\prod_{i=1}^D\pi_i^\star(x^i,y^i).
\]
Its first marginal is $\otimes_i p^i=p$ and its second is $\otimes_iq^i=q$. Each projected pair has mismatch probability $\TV(p^i,q^i)$, so $\Pi^\star$ attains equality in Eq.~\eqref{eq:tensor-lower}. This proves Eq.~\eqref{eq:tensor}.

The product structure ensures that this construction has full marginals $p$ and $q$. For general sequence marginals, the same lower bound holds, and equality requires a full-sequence coupling whose coordinate projections are all maximal couplings. Moreover, the proposition minimizes expected weighted Hamming distance rather than whole-sequence mismatch $\Pr(X\ne Y)$. The minimum of the latter is $\TV(p,q)$, which generally differs from the coordinatewise sum.
\end{proof}

\begin{proof}[Conditional MAFTC exactness]
Fix an endpoint $a$ and abbreviate $p^i=p_t^i(\cdot\mid a^i)$ and $q^i=p_s^i(\cdot\mid a^i)$. Form the product kernel $K^a=\otimes_iK_i^{\rm MA}$, meaning $K^a(y\mid x)=\prod_iK_i^{\rm MA}(y^i\mid x^i)$. For every $y\in\mathcal T^D$, finite-sum factorization gives
\begin{equation}
\begin{aligned}
(p_t(\cdot\mid a)K^a)(y)
&=\sum_{x\in\mathcal T^D}\prod_i p^i(x^i)K_i^{\rm MA}(y^i\mid x^i)\\
&=\prod_i\sum_{x^i\in\mathcal T}p^i(x^i)K_i^{\rm MA}(y^i\mid x^i)\\
&=\prod_iq^i(y^i)=p_s(y\mid a).
\end{aligned}
\end{equation}
Thus the conditional output marginal equals $p_s(\cdot\mid a)$ for any interval length. Each induced coordinate coupling is maximal, so the product coupling attains the lower bound in Proposition~\ref{prop:tensor} for every choice of positive weights. This proves the conditional sequence claim in Section~\ref{sec:sequence}.
\end{proof}

\subsection{Proof of Proposition~\ref{prop:ct} and the excess limit}\label{app:ct-proof}
\begin{proof}
Fix an interior time $t\in(0,1)$ and suppress the subscript. Let $g=[\dot p]_+$ and $\ell=[-\dot p]_+$. Differentiating $\sum_xp_t(x)=1$ gives $\sum_x\dot p(x)=0$, and therefore
\begin{equation}
\sum_xg(x)=\sum_x\ell(x)
=\frac12\sum_x|\dot p(x)|=:M.
\end{equation}
For any admissible flux $j$, define the total inflow and outflow of state $y$ by
\[
I(y)=\sum_{x\ne y}j(y,x),\qquad
O(y)=\sum_{x\ne y}j(x,y).
\]
Our indexing convention is that $j(y,x)$ denotes flux from source $x$ to destination $y$, so the continuity equation is $\dot p(y)=I(y)-O(y)$. On a gain state, $\dot p(y)>0$ and hence
$I(y)=\dot p(y)+O(y)\ge\dot p(y)$. Summing only over gain states gives
\begin{equation}
\Aact(j)=\sum_y I(y)
\ge\sum_{y:\dot p(y)>0}I(y)
\ge\sum_{y:\dot p(y)>0}\dot p(y)=M.
\label{eq:ct-lower}
\end{equation}
For $M>0$, define
\[
j^{\rm MA}(y,x)=\frac{\ell(x)g(y)}{M}.
\]
The supports of $g$ and $\ell$ are disjoint, so the flux is off diagonal. Summing over destinations gives outgoing flux
\[
\sum_{y\ne x}j^{\rm MA}(y,x)
=\ell(x)\frac{\sum_yg(y)}{M}=\ell(x),
\]
and summing over sources gives incoming flux $g(y)$. Hence the induced derivative is $g(y)-\ell(y)=\dot p(y)$ and the total activity is $\sum_x\ell(x)=M$. At the interior time under consideration, $p_t(x)=0$ implies $\dot p_t(x)=0$ because $p_t(x)$ is nonnegative and differentiable. Thus $\ell(x)=0$ on zero-probability states, and the constructed flux satisfies the support condition for admissibility. This attains Eq.~\eqref{eq:ct-lower}. If $M=0$, then $\dot p=0$ and the zero flux is admissible and optimal.

It remains to characterize equality. The first inequality in Eq.~\eqref{eq:ct-lower} is an equality exactly when $I(y)=0$ for every loss or neutral state. The second is an equality exactly when $O(y)=0$ for every gain state, because $I(y)=\dot p(y)+O(y)$. The continuity equation then gives
\[
O(x)=\ell(x)\quad(x\in\{\dot p<0\}),
\qquad
I(y)=g(y)\quad(y\in\{\dot p>0\}),
\]
and all other inflows and outflows vanish. Thus every minimizing flux is supported from the loss set to the gain set and has statewise outflow $\ell$ and inflow $g$. For $M>0$, the array $\gamma(x,y):=j(y,x)/M$ has row marginal $\ell/M$ and column marginal $g/M$. This identifies the minimizing fluxes with the couplings in $\GammaC(\ell/M,g/M)$. The flux in Eq.~\eqref{eq:ct-independent} corresponds to the independent coupling.

Finally, differentiability on the finite state space gives
\[
\left\|\frac{p_{t+h}-p_t}{h}-\dot p_t\right\|_1\longrightarrow0
\qquad(h\downarrow0).
\]
Continuity of the norm then yields
\begin{equation}
\frac{\TV(p_t,p_{t+h})}{h}
=\frac12\left\|\frac{p_{t+h}-p_t}{h}\right\|_1
\longrightarrow\frac12\|\dot p_t\|_1=M.
\end{equation}
To obtain the coupling limit, write $r_h^-=[p_t-p_{t+h}]_+$, $r_h^+=[p_{t+h}-p_t]_+$, and $\tau_h=\TV(p_t,p_{t+h})$. The positive-part map is $1$-Lipschitz, since $|[a]_+-[b]_+|\le|a-b|$. Applying this inequality pointwise gives
\[
\left\|\frac{r_h^-}{h}-\ell\right\|_1\longrightarrow0,\qquad
\left\|\frac{r_h^+}{h}-g\right\|_1\longrightarrow0.
\]
When $M>0$, $\tau_h/h\to M$ ensures that $\tau_h>0$ for sufficiently small $h$, and hence
\[
\frac{r_h^-}{\tau_h}\longrightarrow\frac{\ell}{M},\qquad
\frac{r_h^+}{\tau_h}\longrightarrow\frac{g}{M}
\quad\text{in }\ell^1.
\]
For $x\ne y$, Eq.~\eqref{eq:ma-coupling} now gives
\[
\frac{\pi^{\rm MA}_{t,t+h}(x,y)}{h}
=\frac{[r_h^-(x)/h][r_h^+(y)/h]}{\tau_h/h}
\longrightarrow\frac{\ell(x)g(y)}{M}=j^{\rm MA}(y,x).
\]
Since there are finitely many off-diagonal pairs, the sum of their absolute errors also vanishes:
\[
\sum_{x\ne y}\left|
\frac{\pi^{\rm MA}_{t,t+h}(x,y)}{h}-j^{\rm MA}(y,x)
\right|\longrightarrow0.
\]
This is the claimed $\ell^1$ convergence of the off-diagonal arrays.
If $M=0$, define $j^{\rm MA}=0$. The total off-diagonal mass is
$\tau_h$, so $\sum_{x\ne y}|\pi^{\rm MA}_{t,t+h}(x,y)/h|
=\tau_h/h\to0$. Thus the coupling limit also holds when the
instantaneous minimum vanishes.

\end{proof}

\paragraph{Continuous-time limit of the activity decomposition.}
Let $P_h:=P_{t+h\mid t}$ be the exact transition kernel of a finite-state
CTMC realizing the prescribed path locally, so $p_tP_h=p_{t+h}$.
Assume locally bounded rates and the short-time expansion in
Eq.~\eqref{eq:ctmc-short-time-transition}. With
$j_t(y,x)=p_t(x)u_t(y,x)$ for $x\ne y$, the finite-time definitions give
$\epsilon(P_h)=0$ and $A_{\rm ind}(P_h)=\TV(p_t,p_{t+h})$.
Because there are finitely many states,
\[
\frac{A(P_h)}{h}
=\sum_{x\ne y}p_t(x)\frac{P_h(y\mid x)}{h}
\longrightarrow\sum_{x\ne y}j_t(y,x)=\Aact_t(j).
\]
Subtracting the already proved limit
$\TV(p_t,p_{t+h})/h\to\tfrac12\|\dot p_t\|_1$ establishes
Eq.~\eqref{eq:ct-excess-limit}, including the case $M_t=0$.
Here $A(P_h)$ measures endpoint disagreement over an interval, not its
expected number of jumps. Under the stated local regularity, both have
first-order term $h\Aact_t(j)$; they need not coincide for finite $h$.

\section{Metric-Induced Paths and Localization of Excess Activity}\label{app:kinetic}
We identify the finite-interval loss and gain threshold for a Gibbs path,
then prove the localization of excess activity in kinetic-optimal dynamics. Fix an
endpoint token $a\in\mathcal T$ and a finite nonnegative endpoint-distance
score $\rho(\cdot,a)$. For inverse temperature $\beta\ge0$, let
$Z_a(\beta)$ be the normalizing constant of the conditional path
\begin{equation}
p_\beta(x\mid a)
=\frac{e^{-\beta \rho(x,a)}}{Z_a(\beta)},
\qquad Z_a(\beta):=\sum_{z\in\mathcal T}e^{-\beta\rho(z,a)}.
\label{eq:gibbs-path}
\end{equation}
The calculations below use finite $\beta$, for which every state has
positive probability. We suppress the coordinate index and abbreviate
$p_\beta(\cdot\mid a)$ to $p_\beta$.

For inverse temperatures $0\le\beta_t<\beta_s<\infty$ at two sampling
times, let $\mu(\beta):=\sum_xp_\beta(x)\rho(x,a)$ be the average endpoint
distance under $p_\beta$. The loss
and gain sets are separated by the threshold
\begin{equation}
\rho^*_{t,s}
=
\frac{\log Z_a(\beta_t)-\log Z_a(\beta_s)}
{\beta_s-\beta_t}
=
\frac{1}{\beta_s-\beta_t}
\int_{\beta_t}^{\beta_s}
\mu(\beta)\,d\beta.
\label{eq:gibbs-threshold}
\end{equation}
Lemma~\ref{lem:gibbs-threshold} characterizes the gain and loss sets and proves the shrinking-interval limit
\begin{equation}
\lim_{\beta_s\downarrow\beta_t}\rho^*_{t,s}
=
\mu(\beta_t).
\label{eq:threshold-limit}
\end{equation}

\paragraph{Kinetic activity around the threshold $\mu$.}
At a fixed finite $\beta$, write $\rho_x:=\rho(x,a)$ and
$\mu:=\mu(\beta)=\sum_xp_\beta(x)\rho_x$. Since
$\partial_\beta\log Z_a(\beta)=-\mu$, differentiating the Gibbs path gives
\begin{equation}
\partial_\beta p_\beta(x)=p_\beta(x)(\mu-\rho_x).
\label{eq:gibbs-derivative}
\end{equation}
For the kinetic-optimal construction in \citet[Eq.~(26)]{shaul2025gdp},
also used by FUDOKI \citep{wang2025fudoki}, the flux in our
destination-first convention is therefore
\[
j^{\rm KO}_\beta(y,x)
=\bigl[p_\beta(x)\partial_\beta p_\beta(y)
-p_\beta(y)\partial_\beta p_\beta(x)\bigr]_+
=p_\beta(x)p_\beta(y)[\rho_x-\rho_y]_+,\qquad x\ne y.
\]
For a differentiable nondecreasing schedule with finite $\beta_t$,
returning to model time $t$ multiplies this flux and both activities below
by $\dot\beta_t\ge0$, leaving their decomposition unchanged. Let
$R=\rho(X,a)$ and $R'=\rho(X',a)$ for independent $X,X'\sim p_\beta$,
with $\mu=\mathbb ER$. The activities of the kinetic-optimal and
minimum-activity fluxes are
\begin{equation}
\Aact_{\rm KO}
=\tfrac12\mathbb E|R-R'|,
\qquad
\Aact_{\rm MA}
=\tfrac12\mathbb E|R-\mu|.
\label{eq:kinetic-activities}
\end{equation}
Both fluxes induce the same derivative of the Gibbs probability path.

Theorem~\ref{thm:kinetic} states the localization result in the main text. In the notation above, its within-group identity is equivalently
\begin{equation}
\Aact_{\rm KO}-\Aact_{\rm MA}
=\mathbb E[(R-R')_+\1\{R'>\mu\}]
+\mathbb E[(R-R')_+\1\{R\le\mu\}].
\label{eq:kinetic-excess-expectation}
\end{equation}
The first term is the flux among states whose distances both exceed $\mu$; the second is the flux among states at or below $\mu$. The proof below also establishes the cross-threshold identity.

\begin{lemma}[Threshold structure of Gibbs residuals]\label{lem:gibbs-threshold}
Let $p_\beta(x\mid a)\propto e^{-\beta \rho(x,a)}$ and $0\le\beta_t<\beta_s<\infty$. With $\rho^*_{t,s}$ defined in Eq.~\eqref{eq:gibbs-threshold},
\[
\begin{aligned}
p_{\beta_s}(x\mid a)>p_{\beta_t}(x\mid a)
&\iff \rho(x,a)<\rho^*_{t,s},\\
p_{\beta_s}(x\mid a)=p_{\beta_t}(x\mid a)
&\iff \rho(x,a)=\rho^*_{t,s},\\
p_{\beta_s}(x\mid a)<p_{\beta_t}(x\mid a)
&\iff \rho(x,a)>\rho^*_{t,s}.
\end{aligned}
\]
Moreover, $\rho^*_{t,s}$ is the average of $\mu(\beta)$ over
$[\beta_t,\beta_s]$.
\end{lemma}

\subsection{Proof of Lemma~\ref{lem:gibbs-threshold}}
\begin{proof}
Let $\beta_s>\beta_t$ and write $\Delta\beta=\beta_s-\beta_t$. The ratio of the two Gibbs probabilities satisfies
\begin{equation}
\log\frac{p_{\beta_s}(x\mid a)}{p_{\beta_t}(x\mid a)}
=-\Delta\beta\,\rho(x,a)
+\log Z_a(\beta_t)-\log Z_a(\beta_s).
\label{eq:gibbs-logratio}
\end{equation}
Therefore $p_{\beta_s}(x\mid a)>p_{\beta_t}(x\mid a)$ exactly when
$\rho(x,a)<\rho^*_{t,s}$, equality holds at the threshold, and states with distance above the threshold lose mass. This proves the single-threshold separation in Eq.~\eqref{eq:gibbs-threshold}.

To obtain its integral form, differentiate the finite sum defining the partition function:
\[
\partial_\beta Z_a(\beta)=-\sum_x\rho(x,a)e^{-\beta\rho(x,a)},\qquad
\partial_\beta\log Z_a(\beta)
=\frac{\partial_\beta Z_a(\beta)}{Z_a(\beta)}
=-\mu(\beta).
\]
The fundamental theorem of calculus then gives
\begin{equation}
\log Z_a(\beta_t)-\log Z_a(\beta_s)
=\int_{\beta_t}^{\beta_s}\mu(\beta)\,d\beta.
\label{eq:gibbs-integral}
\end{equation}
Dividing by $\Delta\beta$ yields Eq.~\eqref{eq:gibbs-threshold}.
Continuity of $\mu(\beta)$ on the finite discrete state space gives
$\rho^*_{t,s}\to \mu(\beta_t)$ as $\beta_s\downarrow\beta_t$. By the
residual-coupling characterization in Eq.~\eqref{eq:minimum-activity-family},
every minimum-activity off-diagonal transition moves mass from above the
finite-interval threshold to below it and strictly decreases distance to
$a$.
\end{proof}

For the generative endpoint, assume $\rho(a,a)=0$ and $\rho(x,a)>0$ for $x\ne a$. Then
\[
p_\beta(a\mid a)=\frac{1}{1+\sum_{x\ne a}e^{-\beta\rho(x,a)}}\longrightarrow1
\qquad(\beta\to\infty).
\]
We use this limit to define the terminal marginal $p_\infty(\cdot\mid a)=\delta_a$.

\subsection{Proof of Theorem~\ref{thm:kinetic}}
\begin{proof}
Fix a finite $\beta$ and use $\rho_x:=\rho(x,a)$ and
$\mu:=\sum_xp_\beta(x)\rho_x$. Let $X\sim p_\beta$ and set
$R:=\rho(X,a)$. By Eq.~\eqref{eq:gibbs-derivative}, states below the
average distance gain mass, those above it lose mass, and those at the threshold
have zero probability derivative. Proposition~\ref{prop:ct} therefore
gives the minimum expected jump rate with respect to $\beta$:
\begin{equation}
\Aact_{\rm MA}
=\sum_{\rho_x>\mu}p_\beta(x)(\rho_x-\mu)
=\mathbb E[(R-\mu)_+]
=\tfrac12\mathbb E|R-\mu|,
\label{eq:gibbs-ma}
\end{equation}
where the last equality uses $\mathbb E(R-\mu)=0$.

We next verify that the kinetic-optimal flux induces the same derivative.
For each state $y$, let
$I_{\rm KO}(y):=\sum_{x\ne y}j_\beta^{\rm KO}(y,x)$ and
$O_{\rm KO}(y):=\sum_{x\ne y}j_\beta^{\rm KO}(x,y)$ denote its inflow
and outflow. Their difference is
\[
\begin{aligned}
I_{\rm KO}(y)-O_{\rm KO}(y)
&=p_\beta(y)\sum_xp_\beta(x)
\bigl([\rho_x-\rho_y]_+-[\rho_y-\rho_x]_+\bigr)\\
&=p_\beta(y)\sum_xp_\beta(x)(\rho_x-\rho_y)\\
&=p_\beta(y)(\mu-\rho_y)
=\partial_\beta p_\beta(y),
\end{aligned}
\]
where the identity $[u]_+-[-u]_+=u$ is used in the second line and Eq.~\eqref{eq:gibbs-derivative} in the last. Hence $j_\beta^{\rm KO}$ is an admissible flux for the Gibbs path.

Let $X'\sim p_\beta$ be independent of $X$ and set
$R':=\rho(X',a)$. Regard $X$ as the source and $X'$ as the destination.
Summing the flux over ordered pairs gives
\[
\Aact_{\rm KO}=\mathbb E[(R-R')_+].
\]
Exchangeability implies that $(R-R')_+$ and $(R'-R)_+$ have the same expectation and sum to $|R-R'|$, proving the first identity in Eq.~\eqref{eq:kinetic-activities}.

Now partition the event $R>R'$ into three disjoint classes:
\begin{equation}
\{R>\mu\ge R'\},\qquad
\{R>R'>\mu\},\qquad
\{\mu\ge R>R'\}.
\label{eq:mean-partition}
\end{equation}
The first class is precisely the flux crossing the threshold $\mu$ from its upper side to its lower side. If $R=\mu$ almost surely, both activities vanish and the result is immediate. Otherwise let
$\omega_H=\Pr(R>\mu)$, $\omega_L=\Pr(R\le\mu)$,
$m_+=\mathbb E[R\mid R>\mu]$, and
$m_-=\mathbb E[R\mid R\le\mu]$. Then $\omega_H,\omega_L>0$, $\omega_H+\omega_L=1$, and independence gives
\begin{equation}
\begin{aligned}
\mathbb E[(R-R')\1\{R>\mu\ge R'\}]
&=\omega_H\omega_L(m_+-m_-)\\
&=\omega_H(m_+-\mu)\\
&=\mathbb E[(R-\mu)_+],
\end{aligned}
\label{eq:mean-cross}
\end{equation}
because $\mu=\omega_Hm_++\omega_Lm_-$. Equations~\eqref{eq:gibbs-ma} and \eqref{eq:mean-cross} prove that the total kinetic flux across this threshold equals the minimum activity. The other two classes contribute
\[
\mathbb E[(R-R')_+\1\{R'>\mu\}]
\quad\text{and}\quad
\mathbb E[(R-R')_+\1\{R\le\mu\}],
\]
respectively. Summing the three disjoint contributions from Eq.~\eqref{eq:mean-partition} and subtracting the crossing term proves Eq.~\eqref{eq:kinetic-excess}. The first remainder lies entirely above $\mu$; the second lies at or below $\mu$.
\end{proof}

\section{Conditional and Posterior-Marginalized Updates}\label{app:posterior}
Let $A\sim p_1$ be the data endpoint and fix $0\le t<s\le1$. Suppose $p_t(\cdot\mid a)K^a_{t,s}=p_s(\cdot\mid a)$ for every $a$. For $p_t(x)>0$, the exact endpoint posterior is $p_{1\mid t}(a\mid x)=p_1(a)p_t(x\mid a)/p_t(x)$. On states with $p_t(x)=0$, choose any probability distribution over endpoints. Define
\[
K^\star_{t,s}(y\mid x)=\sum_a p_{1\mid t}(a\mid x)K^a_{t,s}(y\mid x).
\]
The mixture identity $p_t(x)=\sum_a p_1(a)p_t(x\mid a)$ implies that every summand is zero when $p_t(x)=0$. Thus Bayes' identity $p_t(x)p_{1\mid t}(a\mid x)=p_1(a)p_t(x\mid a)$ holds on all input states and yields
\begin{equation}
\begin{aligned}
(p_tK^\star_{t,s})(y)
&=\sum_{x,a}p_1(a)p_t(x\mid a)K^a_{t,s}(y\mid x)\\
&=\sum_a p_1(a)p_s(y\mid a)=p_s(y).
\end{aligned}
\label{eq:posterior-exact}
\end{equation}
For coordinatewise MAFTC, write
$K^a_{t,s}(y\mid x)=\prod_i K^{i,a^i}_{t,s}(y^i\mid x^i)$,
where $K^{i,a^i}_{t,s}$ is the conditional kernel for coordinate $i$.
Algorithm~\ref{alg:maftc} samples endpoint tokens independently from
$q_\theta$, inducing the parallel transition kernel
\begin{equation}
K^\theta_{t,s}(y\mid x)
=\prod_{i=1}^D\sum_{a^i\in\mathcal T}
q_{\theta,i}(a^i\mid x,t)K^{i,a^i}_{t,s}(y^i\mid x^i).
\label{eq:learned-parallel-kernel}
\end{equation}
Appendix~\ref{app:oracle-accuracy} evaluates this parallel construction with
exact token posterior marginals to isolate its finite-step sampling behavior.

\section{Implementation and Numerical Stability}\label{app:implementation}

\subsection{Vectorized MAFTC}
Each stage evaluates the frozen model once,
obtains the learned endpoint posterior at each position,
and samples endpoint tokens independently.
Model-specific conversion and decoding truncation are described in
Appendix~\ref{app:experiments}. The sampled endpoint and the probability path
determine the adjacent marginals $p,q$.

For FUDOKI, the implementation forms normalized log probabilities $\ell_p=\log p$ and $\ell_q=\log q$. At the realized source token $x$ with $p(x)>0$, it computes
\[
\kappa_x=\exp\!\bigl(\min\{0,\ell_q(x)-\ell_p(x)\}\bigr).
\]
For rows selected to move, the gain residual is evaluated on states with $\ell_q(y)>\ell_p(y)$ as
\[
g(y)=e^{\ell_q(y)}\bigl[-\operatorname{expm1}(\ell_p(y)-\ell_q(y))\bigr],
\]
and set to zero elsewhere, where $\operatorname{expm1}(z)=e^z-1$ is evaluated accurately near zero. Categorical sampling normalizes these nonnegative weights. This evaluates $[q-p]_+$ without subtracting nearly equal probabilities. Zero-probability source rows use the identity convention; if $p=q$, every state is retained. At the terminal point-mass target, the update directly returns the sampled endpoint.

For text, let $\alpha\in[0,1]$ parameterize the uniform-source conditional path $p_\alpha(x\mid a)=\alpha\delta_a(x)+(1-\alpha)/V$, and let $\alpha_t$ and $\alpha_s$ denote its values at successive sampling times. For $0\le\alpha_t<\alpha_s\le1$, the gain residual is supported only at $a$. For $x\ne a$, the MAFTC retention probability is
\[
\frac{p_{\alpha_s}(x\mid a)}{p_{\alpha_t}(x\mid a)}
=\frac{(1-\alpha_s)/V}{(1-\alpha_t)/V}
=\frac{1-\alpha_s}{1-\alpha_t}.
\]
A move therefore goes directly to $a$, while a token already equal to $a$ remains fixed. Equal interpolation parameters give the identity update.

\subsection{Complexity}
For batch size $B$, sequence length $D$, and vocabulary size $V$, dense MAFTC uses $O(BDV)$ arithmetic and storage per stage, including the residual tensors, without constructing a $V\times V$ coupling matrix. The uniform-source conditional update reduces to $O(BD)$ once endpoint tokens have been sampled. Both implementations use one neural evaluation per stage.

\section{Experimental Details and Full Results}\label{app:experiments}
\raggedbottom
\subsection{Evaluation quantities}\label{app:metric-definitions}
We report sample averages of $L_{\rm act}$, $D_{\rm net}$, and $R_{\rm traj}$ from Eq.~\eqref{eq:trajectory-excess}. Complete trajectory statistics are included in the accompanying result files.

For each prescribed conditional pair $(p,q)$ and Markov kernel $K$ on the discrete state space, we evaluate
\[
\widetilde q(y)=\sum_x p(x)K(y\mid x),\qquad
A(K)=1-\sum_x p(x)K(x\mid x).
\]
The resulting $A_{\rm pres}$, $A_{\rm ind}$, $\chi$, and $\epsilon$ are defined in Eqs.~\eqref{eq:activity-coordinates}--\eqref{eq:defect-decomposition}. We sum activity and its components over intervals and average over endpoints. Marginal mismatch is reported as the average over both endpoints and intervals, denoted $\bar\epsilon$. Each conditional evaluation uses the prescribed input distribution $p$, so $\bar\epsilon$ measures one-step marginal mismatch under prescribed input marginals. The conditional activity-to-minimum ratio divides summed expected activity by the summed prescribed minimum. Terminal sample quality is evaluated separately on full rollouts.

For text, the external evaluator $p_{\rm eval}$ is GPT-2 large \citep{radford2019language}.\footnote{\url{https://huggingface.co/openai-community/gpt2-large}} We decode generated tokens with the GPT-2 tokenizer, remove special tokens, and retokenize with the evaluator's tokenizer without adding special tokens. Each sequence is truncated to at most 1,024 evaluator tokens. For a retokenized sequence $z_j$ of length $L_j$, we score positions $2,\ldots,L_j$ conditioned on their prefixes, excluding padding. We report
\[
\operatorname{Gen\text{-}PPL}
=\exp\!\left[-\frac{1}{M}\sum_j\sum_{m=2}^{L_j}
\log p_{\rm eval}(z_{j,m}\mid z_{j,<m})\right],
\qquad M=\sum_j(L_j-1).
\]
Here $M$ is the total number of scored tokens, so corpus negative log-likelihood is token-weighted rather than averaged equally across sequences. Lexical metrics use the original generated token IDs with padding removed and EOS retained. We distinguish corpus-level from within-sequence diversity, following the set-level and string-level distinction in \citet{wiher2022decoding}. Pooled unigram entropy is $-\sum_{v=1}^V\widehat f_v\log\widehat f_v$, where $\widehat f_v$ is the empirical relative token frequency across all sequences. We also report unigram entropy averaged over individual sequences. We use natural logarithms and the convention $0\log0=0$. Following the Distinct family of measures \citep{li2016diversity}, pooled Dist-$n$ divides the number of distinct token $n$-grams across the corpus by the number of within-sequence $n$-gram positions; no $n$-grams cross sequence boundaries. For each sequence, Rep-4 is one minus the number of distinct four-grams divided by the number of four-gram positions; we report its average over sequences. Since Distinct scores depend on text length \citep{liu2022rethinking}, we compare corpora with the same number of generated sequences and the same prescribed sequence length.

\paragraph{Uncertainty and comparison units.}
We use 10,000 bootstrap draws. Text resampling uses complete sequences and
recomputes corpus Gen-PPL from total negative log-likelihood and scored token
counts. Image resampling uses prompt clusters, retaining all four images and
stratifying by the six GenEval tasks, whose average accuracies retain equal weights.
Paired comparisons share the resampled indices across methods. We report
95\% percentile bootstrap confidence intervals (CIs). Unless an adjustment is
stated, each interval applies to an individual quantity or comparison
(pointwise coverage). An interval
containing zero does not establish equivalence, and these fixed-seed intervals
do not estimate variation across training or generation seeds.

Figure~\ref{fig:text-entropy} uses marginal intervals for the plotted Gen-PPL and
average sequence entropy; their overlap is not a test of a paired difference.
Pooled entropy and pooled Dist-4 are reported as corpus-level descriptive statistics.
The full-state conditional-kernel calculations are deterministic for the chosen
endpoints. Oracle sweep bands show the interquartile range across
target/geometry seeds.

\subsection{Full-state evaluation of conditional kernels}\label{app:conditional-evaluation}
For FUDOKI, we select 32 endpoint token IDs uniformly without replacement from the 16,384-entry codebook, using a PyTorch CPU random permutation with seed 1208. We hold this set fixed across methods and use uniform grids with 8, 16, or 32 steps. For endpoint $a$ and interval $[t_k,t_{k+1}]$, the prescribed pair is $p=p_{t_k}(\cdot\mid a)$ and $q=p_{t_{k+1}}(\cdot\mid a)$. The path uses $\beta_t=3(t/(1-t))^{0.9}$ and squared cosine-distance geometry $\rho(x,a)=(2-2\langle\bar e_x,\bar e_a\rangle)^2$, where $\bar e$ denotes a normalized codebook embedding. We evaluate each implemented conditional kernel over the full state space in double precision using the original codebook distances. This directly yields the induced output marginal and the activity decomposition, without sampled histograms. All methods use the same fixed endpoint token for each conditional pair and reveal it directly in the final interval.

\begin{table}[ht]
\caption{\textbf{FUDOKI conditional activity decomposition and marginal mismatch.} Activity columns sum over intervals and average over 32 endpoints. The final column averages one-step marginal mismatch in total variation over both intervals and endpoints. Thus $\sum_k A_k=\sum_k A_{{\rm ind},k}+\sum_k\chi_k$, while $\bar\epsilon$ is measured separately. MAFTC's zero entries are rounded numerical values.}
\label{tab:conditional-audit}
\centering
\begin{tabular}{c l rrrrr}
\toprule
NFE & Method & $\sum_k A_{{\rm pres},k}$ & $\sum_k A_k$ & $\sum_k A_{{\rm ind},k}$ & $\sum_k\chi_k$ & $\bar\epsilon$\\
\midrule
8 & Euler & 2.8126 & 2.2937 & 1.9113 & \underline{0.3824} & 0.11830\\
8 & TC    & 2.8126 & 3.1372 & 2.5497 & 0.5875 & \underline{0.03775}\\
8 & MAFTC & 2.8126 & 2.8126 & 2.8126 & \textbf{0.0000} & \textbf{0.00000}\\
\midrule
16 & Euler & 2.9849 & 2.8166 & 2.2531 & \underline{0.5635} & 0.04810\\
16 & TC    & 2.9849 & 3.4888 & 2.7557 & 0.7331 & \underline{0.01595}\\
16 & MAFTC & 2.9849 & 2.9849 & 2.9849 & \textbf{0.0000} & \textbf{0.00000}\\
\midrule
32 & Euler & 3.0186 & 3.0314 & 2.2823 & \underline{0.7491} & 0.02419\\
32 & TC    & 3.0186 & 3.6474 & 2.7613 & 0.8860 & \underline{0.00865}\\
32 & MAFTC & 3.0186 & 3.0186 & 3.0186 & \textbf{0.0000} & \textbf{0.00000}\\
\bottomrule
\end{tabular}
\end{table}

At 32 NFE, Euler's summed activity is 3.031, close to the prescribed minimum of 3.019. Its induced marginal transport requires only 2.282 changes, leaving 0.749 in coupling excess. The excess nearly offsets the gap between the induced and prescribed minima: total activity alone obscures the different contributions. TC reduces one-step marginal mismatch relative to Euler at all three evaluated NFEs, but has greater coupling excess.

For Duo, we use the uniform-source conditional path and closed-form ancestral transition specified below. Duo+DCD's conditional activity-to-minimum ratios are 2.58590, 3.29904, and 3.98325 at 8, 16, and 32 NFE, respectively; MAFTC's are one, and both kernels realize $pK=q$.

\subsection{Integrated expected jump rate}\label{app:integrated-activity}
For the same 32 FUDOKI endpoints, we integrate the minimum and kinetic-optimal activities in Eq.~\eqref{eq:kinetic-activities} along each full conditional path, with inverse temperature $\beta$ ranging from zero to infinity. Their endpoint-averaged integrals are 3.0381 and 4.0124, respectively. By Theorem~\ref{thm:kinetic}, the difference, 0.9743, is the integrated within-group flux. Its ratio to the integrated kinetic-optimal activity is 24.3\%, computed from endpoint-averaged integrals. We use 2,048-interval composite Simpson quadrature in $\log(1+\beta)$ up to $\beta_{\max}\approx1964.98$; the remaining tail is bounded by $6.2\times10^{-15}$.

To bound the omitted tail, fix an endpoint $a$, let $m_a:=\min_x\rho(x,a)$, and write $\widetilde\rho_x:=\rho(x,a)-m_a$ and $\widetilde Z_a(\beta):=\sum_xe^{-\beta \widetilde\rho_x}$. Subtracting $m_a$ leaves the Gibbs probabilities unchanged. For $X,X'\sim p_\beta$ independently, nonnegativity gives
\[
\Aact_{\rm KO}(\beta)=\tfrac12\mathbb E|\widetilde\rho_X-\widetilde\rho_{X'}|
\le\mathbb E \widetilde\rho_X,\qquad
\Aact_{\rm MA}(\beta)=\tfrac12\mathbb E|\widetilde\rho_X-\mathbb E \widetilde\rho_X|
\le\mathbb E \widetilde\rho_X.
\]
The evaluated paths have a unique minimum-distance state, so $\widetilde Z_a(\beta)\to1$ as $\beta\to\infty$. Since $-\partial_\beta\log\widetilde Z_a(\beta)=\mathbb E \widetilde\rho_X$, either activity satisfies
\[
\int_{\beta_{\max}}^\infty\Aact_{\bullet}(\beta)\,d\beta
\le\int_{\beta_{\max}}^\infty\mathbb E \widetilde\rho_X\,d\beta
=\log\widetilde Z_a(\beta_{\max}),\qquad
\bullet\in\{\mathrm{MA},\mathrm{KO}\}.
\]
Across the 32 endpoints, the maximum omitted-tail bound is $6.16\times10^{-15}$. The maximum change in the retained-interval integral between the final two nested Simpson grids is $3.26\times10^{-6}$, which we use to assess numerical convergence.

\subsection{Finite-step baseline definitions}\label{app:baseline-definitions}
The Euler sampler in the released FUDOKI implementation \citep{wang2025fudoki} uses a coordinatewise exponential update with frozen rates. After drawing endpoint token $a$, it freezes the outgoing kinetic-optimal rates $u_{t_k}(\cdot,x\mid a)$ over a step of width $h=t_{k+1}-t_k$ and samples at most one coordinate jump, with total outgoing rate $\lambda_k$:
\begin{align*}
\Pr(Y\ne x\mid x,a)&=1-e^{-h\lambda_k},
&\lambda_k&=\sum_{z\ne x}u_{t_k}(z,x\mid a),\\
\Pr(Y=y\mid Y\ne x,x,a)&=\frac{u_{t_k}(y,x\mid a)}{\lambda_k}.
\end{align*}
The destination distribution is used only when $\lambda_k>0$; if $\lambda_k=0$, the update retains $x$.
The time-corrected (TC) sampler uses the same endpoint-conditioned rate geometry but integrates its schedule dependence across the interval while keeping the source state fixed \citep{wan2026corrected}. Our implementation uses four left-endpoint quadrature nodes to approximate the first-jump distribution, with at most one jump per coordinate per step. Euler and TC both use the implementation's convention $\dot\beta_0=0$. TC corrects the frozen-time approximation without an additional model evaluation.

For Duo+DCD \citep{sahoo2025duality}, we use the uniform-source conditional
path $p_\alpha(z\mid a)=\alpha\delta_a(z)+(1-\alpha)/V$.
Let $0\le\alpha_k<\alpha_{k+1}\le1$ denote adjacent interpolation values
and set $r_k=\alpha_k/\alpha_{k+1}$. For one coordinate, the conditional
ancestral bridge is
\begin{equation}
B_k^a(y\mid x)
=\frac{p_{\alpha_{k+1}}(y\mid a)}{p_{\alpha_k}(x\mid a)}
\left[r_k\1\{x=y\}+\frac{1-r_k}{V}\right].
\label{eq:duo-conditional-bridge}
\end{equation}
For the full current sequence $x$, the update averages this bridge over
the learned endpoint posterior at each coordinate:
\[
K_k^{\mathrm{Duo+DCD},i}(y^i\mid x)
=
\sum_{a^i\in\mathcal T} q_{\theta,i}(a^i\mid x,\alpha_k)
B_k^{a^i}(y^i\mid x^i).
\]
We parameterize model predictions by the interpolation value $\alpha_k$
at time $t_k$; Appendix~\ref{app:duo-language} specifies its grid.
For both Duo+DCD and MAFTC, the endpoint probabilities are obtained from the truncated denoiser probabilities $b_{\theta,i}$ by
\[
q_{\theta,i}(a^i\mid x,\alpha)\propto b_{\theta,i}(a^i\mid x,\alpha)
\left[\alpha\1\{a^i=x^i\}+\frac{1-\alpha}{V}\right],
\]
with normalization over $a^i\in\mathcal T$. This is the learned endpoint posterior
for the $i$-th token used in Algorithm~\ref{alg:maftc}.

For the $\Psi$-sampler baseline, we use the posterior mixture of \citet[Eq.~(12)]{deschenaux2026psi}. With $\omega_k$ denoting the weight on the ancestral kernel, the update for coordinate $i$ is
\[
K_k^{\Psi,i}(y^i\mid x)=\omega_k K_k^{\mathrm{Duo+DCD},i}(y^i\mid x)
+(1-\omega_k)\left[\alpha_{k+1}q_{\theta,i}(y^i\mid x,\alpha_k)
+\frac{1-\alpha_{k+1}}{V}\right].
\]
We fix the implementation's rescale parameter to $\eta=0.05$. For nonterminal steps, an auxiliary correction parameter $\sigma_k$ determines the mixture weight $\omega_k$ through
\[
\sigma_k=\eta\min\!\left\{\eta,\frac{1-\alpha_{k+1}}{\alpha_k}\right\},
\qquad
\omega_k=\max\!\left\{0,\min\!\left\{1,1-\frac{\sigma_k}{1-\alpha_{k+1}}\right\}\right\}.
\]
The final step uses $\omega_{N-1}=1$. All methods use the same $\alpha$ grid specified below and one model prediction per step; no $\Psi$ hyperparameters are tuned for this evaluation.

\subsection{FUDOKI}
We use the released \texttt{LucasJinWang/FUDOKI} checkpoint,\footnote{\url{https://huggingface.co/LucasJinWang/FUDOKI}} whose transformer has 24 layers, hidden width 2,048, and 16 attention heads. Its vector-quantized (VQ) image tokenizer, VQ-16, downsamples each spatial dimension by a factor of 16 and has 16,384 codes with eight-dimensional embeddings. We generate a $24\times24$ latent grid (576 tokens), decoded to $384\times384$ pixels. The model and tokenizer remain frozen. Image tokens are initialized independently from the uniform distribution, and endpoint predictions use the full vocabulary without top-$k$ or nucleus truncation.

Sampling uses $t_k=k/N$ for $k=0,\ldots,N$, with the Gibbs path and $\beta_t$ schedule specified in Appendix~\ref{app:conditional-evaluation}. We retain FUDOKI's classifier-free guidance (CFG) convention, $\ell_{\rm cfg}=(1+w)\ell_{\rm cond}-w\ell_{\rm uncond}$, where $\ell_{\rm cond}$ and $\ell_{\rm uncond}$ are the conditional and unconditional logits and $w=5$ is the guidance scale. Reported NFE counts calls to this guided prediction function, so $N$ sampling steps require $N$ guided predictions and $2N$ backbone forward passes for every image sampler.

For each sampler and NFE, we evaluate all 553 GenEval prompts with four images per prompt and generation seed 3407, recording every image's complete token trajectory. GenEval is the macro average of its six task accuracies. Table~\ref{tab:full-primary} also reports average raw image--text cosine similarity from CLIP ViT-B/32 with OpenAI weights \citep{radford2021clip}, evaluated over all 2,212 images, and the activity quantities from Appendix~\ref{app:metric-definitions}.

\begin{table}[ht]
\caption{\textbf{Complete FUDOKI benchmark.} GenEval and CLIP cosine similarity are higher-is-better; $L_{\rm act}$ and $R_{\rm traj}$ are average cumulative token changes and trajectory excess. Each configuration contains 2,212 images from 553 prompts and complete per-image trajectories.}
\label{tab:full-primary}
\centering
\small
\setlength{\tabcolsep}{5pt}
\begin{tabular}{c l r r r r}
\toprule
NFE & Sampler & GenEval $\uparrow$ & CLIP $\uparrow$ & $L_{\rm act}$ & $R_{\rm traj}$ \\
\midrule
4 & Euler & 0.5106 & 0.3132 & \textbf{1486.0} & \textbf{910.1} \\
 & TC & \textbf{0.5958} & \textbf{0.3245} & 1844.3 & 1268.4 \\
 & MAFTC & \underline{0.5872} & \underline{0.3222} & \underline{1671.8} & \underline{1095.9} \\
\addlinespace[2pt]
8 & Euler & 0.6903 & 0.3328 & \underline{2146.3} & \underline{1570.5} \\
 & TC & \textbf{0.7221} & \textbf{0.3354} & 2276.8 & 1701.0 \\
 & MAFTC & \underline{0.7122} & \underline{0.3337} & \textbf{1879.6} & \textbf{1304.2} \\
\addlinespace[2pt]
16 & Euler & \underline{0.7508} & \underline{0.3372} & \underline{2307.4} & \underline{1732.0} \\
 & TC & \textbf{0.7572} & \textbf{0.3380} & 2376.3 & 1801.0 \\
 & MAFTC & 0.7429 & 0.3362 & \textbf{1893.9} & \textbf{1320.0} \\
\addlinespace[2pt]
32 & Euler & \textbf{0.7531} & \textbf{0.3382} & \underline{2407.8} & \underline{1833.2} \\
 & TC & \underline{0.7453} & \underline{0.3376} & 2450.4 & 1876.1 \\
 & MAFTC & 0.7443 & 0.3371 & \textbf{1882.7} & \textbf{1310.6} \\
\addlinespace[2pt]
64 & Euler & \textbf{0.7528} & \textbf{0.3367} & \underline{2492.0} & \underline{1917.9} \\
 & TC & \underline{0.7452} & \underline{0.3366} & 2529.6 & 1955.4 \\
 & MAFTC & 0.7438 & 0.3357 & \textbf{1956.1} & \textbf{1384.4} \\
\addlinespace[2pt]
128 & Euler & 0.7268 & \underline{0.3349} & \underline{2618.0} & \underline{2044.0} \\
 & TC & \textbf{0.7392} & \textbf{0.3351} & 2648.5 & 2074.5 \\
 & MAFTC & \underline{0.7369} & 0.3344 & \textbf{2067.0} & \textbf{1495.3} \\
\bottomrule
\end{tabular}
\end{table}

Table~\ref{tab:geneval-categories} reports the six task accuracies underlying GenEval. At NFE 4, MAFTC improves all six categories over Euler. The category trends vary on finer grids: from NFE 64 to 128, MAFTC's color-attribute and counting accuracies increase, while its position and color accuracies decrease.

\begin{table}[ht]
\caption{\textbf{Complete FUDOKI GenEval task breakdown.} All accuracies are higher-is-better. Overall is the macro average of the six task accuracies. Each configuration contains 2,212 images from all 553 prompts under the protocol of Table~\ref{tab:full-primary}. MAFTC rows are shaded.}
\label{tab:geneval-categories}
\centering
\fontsize{9}{10.5}\selectfont
\setlength{\tabcolsep}{3pt}
\begin{tabular}{clrrrrrrr}
\toprule
NFE & Sampler & Overall & \shortstack[r]{Single\\obj.} & \shortstack[r]{Two\\obj.} & Counting & Colors & Position & \shortstack[r]{Color\\attr.} \\
\midrule
4 & Euler & 0.5106 & 0.8563 & 0.5404 & 0.3281 & \underline{0.7314} & 0.4425 & \underline{0.1650} \\
 & TC & \textbf{0.5958} & \textbf{0.9156} & \textbf{0.6364} & \underline{0.3906} & \textbf{0.7846} & \textbf{0.5500} & \textbf{0.2975} \\
\rowcolor{gray!10}
 & MAFTC & \underline{0.5872} & \underline{0.8750} & \underline{0.6263} & \textbf{0.4125} & \textbf{0.7846} & \underline{0.5275} & \textbf{0.2975} \\
\addlinespace[2pt]
8 & Euler & 0.6903 & \underline{0.9406} & 0.7727 & 0.4500 & 0.8537 & \underline{0.6200} & 0.5050 \\
 & TC & \textbf{0.7221} & \textbf{0.9438} & \textbf{0.8005} & \textbf{0.5125} & \underline{0.8883} & \textbf{0.6525} & \underline{0.5350} \\
\rowcolor{gray!10}
 & MAFTC & \underline{0.7122} & 0.9219 & \underline{0.7854} & \underline{0.4719} & \textbf{0.9016} & \textbf{0.6525} & \textbf{0.5400} \\
\addlinespace[2pt]
16 & Euler & \underline{0.7508} & \textbf{0.9594} & \underline{0.8308} & 0.4875 & \textbf{0.9096} & \textbf{0.7050} & \underline{0.6125} \\
 & TC & \textbf{0.7572} & \underline{0.9531} & \textbf{0.8788} & \underline{0.4969} & \textbf{0.9096} & 0.6750 & \textbf{0.6300} \\
\rowcolor{gray!10}
 & MAFTC & 0.7429 & 0.9438 & 0.8056 & \textbf{0.5406} & \underline{0.8750} & \underline{0.6925} & 0.6000 \\
\addlinespace[2pt]
32 & Euler & \textbf{0.7531} & \textbf{0.9594} & \underline{0.8384} & \textbf{0.4969} & \underline{0.9016} & \underline{0.6650} & \underline{0.6575} \\
 & TC & \underline{0.7453} & \underline{0.9500} & 0.8333 & 0.4594 & \textbf{0.9043} & 0.6600 & \textbf{0.6650} \\
\rowcolor{gray!10}
 & MAFTC & 0.7443 & \underline{0.9500} & \textbf{0.8561} & \underline{0.4844} & 0.8803 & \textbf{0.6725} & 0.6225 \\
\addlinespace[2pt]
64 & Euler & \textbf{0.7528} & \underline{0.9250} & 0.8460 & \underline{0.4844} & \textbf{0.8963} & \textbf{0.6700} & \textbf{0.6950} \\
 & TC & \underline{0.7452} & \textbf{0.9313} & \textbf{0.8561} & 0.4813 & \underline{0.8777} & \underline{0.6575} & \underline{0.6675} \\
\rowcolor{gray!10}
 & MAFTC & 0.7438 & \underline{0.9250} & \underline{0.8510} & \textbf{0.5000} & 0.8670 & \textbf{0.6700} & 0.6500 \\
\addlinespace[2pt]
128 & Euler & 0.7268 & \textbf{0.9250} & \underline{0.8283} & 0.4531 & \textbf{0.8697} & \underline{0.6425} & \underline{0.6425} \\
 & TC & \textbf{0.7392} & \underline{0.9125} & 0.8258 & \textbf{0.5250} & \underline{0.8670} & \textbf{0.6925} & 0.6125 \\
\rowcolor{gray!10}
 & MAFTC & \underline{0.7369} & 0.9063 & \textbf{0.8434} & \underline{0.5094} & 0.8324 & 0.6375 & \textbf{0.6925} \\
\bottomrule
\end{tabular}
\end{table}

\subsubsection{Paired uncertainty and visual examples}
\begin{figure}[p]
\centering
\includegraphics[width=\linewidth]{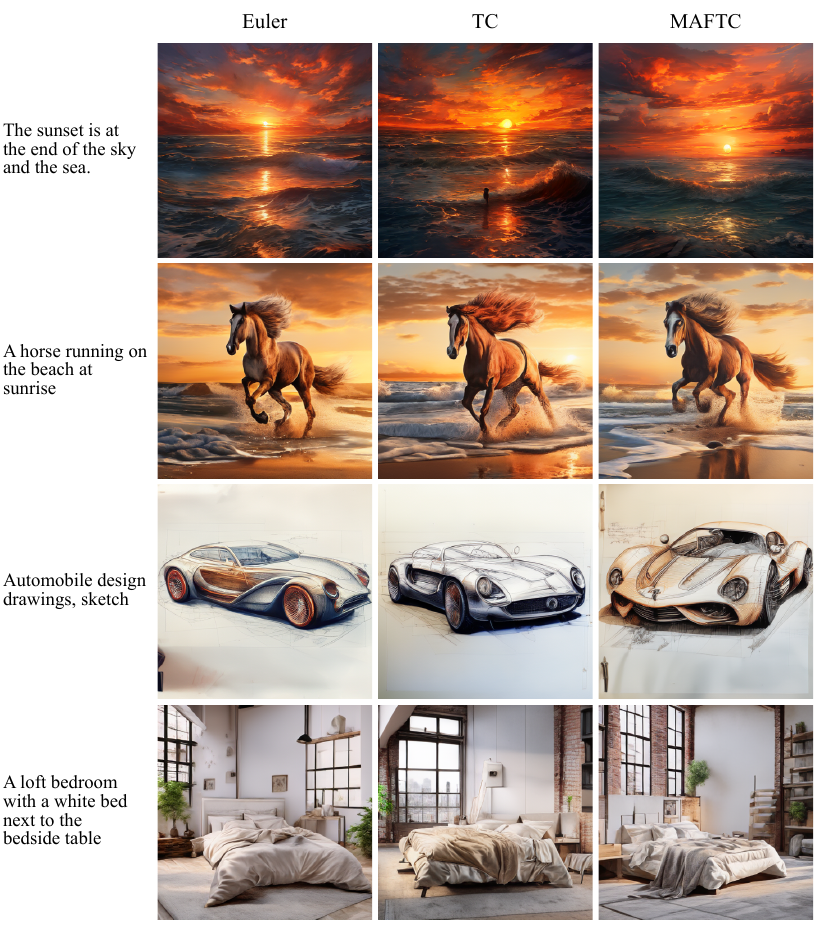}
\caption{\textbf{Image generation on varied prompts at 32 NFE.}
Euler, TC, and MAFTC use the same frozen FUDOKI checkpoint, CFG scale 5,
base seed 3407, and $384\times384$ resolution. For each prompt, all three
samplers start from identical image tokens. Prompts and sample index zero
were fixed before generation; all twelve outputs are shown.}
\label{fig:image-rich-scenes}
\end{figure}

Table~\ref{tab:paired-summary}(a) reports both baseline comparisons at the
budgets in Table~\ref{tab:generation-quality}, using all 553 prompts. The complete
4--128 NFE interval grid is in \texttt{supplementary\_tables.zip}.
MAFTC improves GenEval over Euler at 4 NFE by $7.66$ percentage points
$[5.73,9.66]$ and at 8 NFE by $2.19$ points $[0.34,4.07]$; all intervals
for its differences from TC include zero. At 4 NFE, it also makes $185.8$
$[179.5,191.9]$ more changes than Euler, illustrating why activity and
terminal quality are evaluated together.
Figure~\ref{fig:image-rich-scenes} compares Euler, TC, and MAFTC on four
prompts from the FUDOKI project page, covering a landscape, an animal,
a drawing, and an interior scene.\footnote{\url{https://fudoki-hku.github.io/}}

\subsubsection{Human preference evaluation}\label{app:hpsv3}
We complement GenEval and CLIP with HPSv3 \citep{ma2025hpsv3}, a
successor to HPSv2 \citep{wu2023hpsv2} that predicts human preferences
from image--prompt pairs. We score the same FUDOKI outputs as
Table~\ref{tab:full-primary} at 4, 8, 16, 32, 64, and 128 NFE: all 553
prompts with four images each, giving 39,816 images across the 18
configurations. We use the official HPSv3 weights and preprocessing,
scoring each image individually. Confidence intervals use 20,000 paired
bootstrap resamples of prompts, retaining all four images for each
sampler within a prompt.

\begin{table}[ht]
\caption{\textbf{HPSv3 on GenEval prompts.} Higher is better. Each score averages all 2,212 images for the corresponding sampler and NFE. The last two columns report paired differences between MAFTC and each baseline, with 95\% bootstrap confidence intervals. Bold and underlining indicate the highest and second-highest scores at each NFE.}
\label{tab:hpsv3}
\centering
\small
\setlength{\tabcolsep}{4pt}
\begin{tabular}{rrrrrr}
\toprule
 & \multicolumn{3}{c}{HPSv3 $\uparrow$} & \multicolumn{2}{c}{MAFTC minus baseline [95\% CI]} \\
\cmidrule(lr){2-4}\cmidrule(lr){5-6}
NFE & Euler & TC & MAFTC & \multicolumn{1}{c}{Euler} & \multicolumn{1}{c}{TC} \\
\midrule
4 & -1.214 & \textbf{1.998} & \underline{1.268} & $+2.482\;[+2.257, +2.714]$ & $-0.731\;[-0.930, -0.530]$ \\
8 & 4.231 & \textbf{5.235} & \underline{4.648} & $+0.417\;[+0.271, +0.567]$ & $-0.587\;[-0.735, -0.441]$ \\
16 & \underline{5.890} & \textbf{6.280} & 5.784 & $-0.106\;[-0.249, +0.038]$ & $-0.497\;[-0.650, -0.346]$ \\
32 & \underline{6.358} & \textbf{6.372} & 6.195 & $-0.163\;[-0.295, -0.035]$ & $-0.177\;[-0.311, -0.043]$ \\
64 & \underline{6.389} & \textbf{6.401} & 6.174 & $-0.215\;[-0.361, -0.072]$ & $-0.227\;[-0.365, -0.091]$ \\
128 & \textbf{6.235} & 6.121 & \underline{6.144} & $-0.091\;[-0.242, +0.058]$ & $+0.023\;[-0.117, +0.163]$ \\
\bottomrule
\end{tabular}
\end{table}

Table~\ref{tab:hpsv3} shows that MAFTC scores above Euler and below TC
at 4 and 8 NFE. At 16--64 NFE, its score estimates are lower than both
baselines, alongside fewer token changes (Table~\ref{tab:full-primary}).
At 128 NFE, MAFTC's score lies between those of the two baselines, and
both paired difference intervals include zero.

\FloatBarrier
\subsubsection{Results at a lower guidance scale}\label{app:fudoki-lower-guidance}
{\raggedright\noindent
\textbf{Image generation quality and token changes at $w=3$.}
We use the same FUDOKI checkpoint and 553 GenEval prompts as in the $w=5$ comparison.
At each NFE, all three samplers start from the same four initial token grids for each prompt.\par}

\begin{center}
\small
\setlength{\tabcolsep}{4pt}
\begin{tabular}{r rrr rrr rrr}
\toprule
 & \multicolumn{3}{c}{GenEval $\uparrow$} & \multicolumn{3}{c}{HPSv3 $\uparrow$} & \multicolumn{3}{c}{NTC $\downarrow$} \\
\cmidrule(lr){2-4}\cmidrule(lr){5-7}\cmidrule(lr){8-10}
NFE & Euler & TC & MAFTC & Euler & TC & MAFTC & Euler & TC & MAFTC \\
\midrule
4 & 0.5650 & \textbf{0.6643} & \underline{0.6313} & 0.438 & \textbf{3.469} & \underline{2.541} & \textbf{1461} & 1785 & \underline{1589} \\
8 & 0.7093 & \textbf{0.7266} & \underline{0.7162} & 4.920 & \textbf{5.837} & \underline{5.269} & \underline{2022} & 2160 & \textbf{1783} \\
16 & 0.7499 & \textbf{0.7604} & \underline{0.7544} & 6.514 & \textbf{6.797} & \underline{6.532} & \underline{2219} & 2297 & \textbf{1837} \\
32 & \textbf{0.7741} & \underline{0.7633} & 0.7604 & \underline{7.022} & \textbf{7.158} & 6.996 & \underline{2344} & 2396 & \textbf{1836} \\
64 & \underline{0.7633} & \textbf{0.7670} & 0.7632 & \underline{7.163} & \textbf{7.205} & 7.109 & \underline{2428} & 2453 & \textbf{1874} \\
128 & \underline{0.7611} & 0.7560 & \textbf{0.7638} & \underline{7.204} & \textbf{7.218} & 7.200 & \underline{2485} & 2501 & \textbf{1920} \\
\bottomrule
\end{tabular}
\end{center}

As the table shows, TC has the highest HPSv3 score at all six budgets.
The gap between the reported TC and MAFTC scores narrows at each successive budget,
from 0.928 points at 4 NFE to 0.018 at 128 NFE.

\FloatBarrier
\subsection{Language generation with Duo}\label{app:duo-language}
We use \texttt{s-sahoo/duo-distilled},\footnote{\url{https://huggingface.co/s-sahoo/duo-distilled}} the released Duo checkpoint produced by DCD. It has 12 transformer blocks, hidden width 768, 12 attention heads, and a 1,024-token context. Its vocabulary contains the 50,257 GPT-2 byte-pair tokens and one padding token ($V=50,258$). OpenWebText is the checkpoint's training corpus; our evaluation consists of unconditional generations from independent uniform token strings. Each cell contains 2,048 sequences of length 1,024, using seed 3407 and shared initial states across samplers. Nucleus sampling \citep{holtzman2020degeneration} with top-$p=0.9$ is applied to the denoiser probabilities before the endpoint-posterior conversion in Appendix~\ref{app:baseline-definitions}.

The conditional probability path and sampling grid are
\[
p_\alpha(x\mid a)=\alpha\1\{x=a\}+\frac{1-\alpha}{V},\qquad
\alpha_k=\delta+\frac{k}{N-1}\bigl[1-\delta(1-\delta)-\delta\bigr]
\quad(0\le k<N),
\]
with $\delta=10^{-3}$ and a final target $\alpha_N=1$. The model is queried at the $N$ equally spaced entrance values from $\delta$ to $1-\delta(1-\delta)$, using its noise input $-\log\alpha_k$; the last query supplies the terminal reveal. We initialize from uniform token strings and make the first model query at $\alpha_0=\delta$. We use $N\in\{2,4,8,16,32,64,128\}$, so this protocol performs exactly $N$ neural evaluations, each requiring one backbone forward pass. Table~\ref{tab:language-quality} reports generative perplexity and lexical metrics for all three samplers.
\begin{table}[H]
\caption{\textbf{Full text quality and token-diversity results.} $H_{\rm pool}$ and $H_{\rm seq}$ are pooled and average sequence unigram entropy (nats), $D_n$ is pooled Dist-$n$, and $R_4$ is average sequence Rep-4. Each cell contains 2,048 length-1,024 sequences at seed 3407. Lexical metrics exclude PAD but retain EOS.}
\label{tab:language-quality}
\centering
\setlength{\tabcolsep}{3.2pt}
\small
\begin{tabular}{c l rrrrrrr}
\toprule
NFE & Method & Gen-PPL $\downarrow$ & $H_{\rm pool}$ & $H_{\rm seq}$ & $D_1$ & $D_2$ & $D_4$ & $R_4$\\
\midrule
2 & Duo+DCD & 924.20 & 6.48 & 5.39 & 0.00963 & 0.3752 & 0.9518 & 0.0018\\
 & $\Psi$ & \textbf{922.05} & 6.48 & 5.38 & 0.00963 & 0.3748 & 0.9519 & 0.0017\\
 & MAFTC & \underline{923.96} & 6.48 & 5.39 & 0.00963 & 0.3751 & 0.9517 & 0.0018\\
\addlinespace[2pt]
4 & Duo+DCD & \underline{156.00} & 6.39 & 5.21 & 0.01613 & 0.2639 & 0.8897 & 0.0051\\
 & $\Psi$ & \textbf{155.75} & 6.39 & 5.21 & 0.01606 & 0.2632 & 0.8894 & 0.0052\\
 & MAFTC & 158.25 & 6.41 & 5.23 & 0.01625 & 0.2674 & 0.8915 & 0.0049\\
\addlinespace[2pt]
8 & Duo+DCD & \underline{59.65} & 6.83 & 5.38 & 0.01857 & 0.2771 & 0.8624 & 0.0104\\
 & $\Psi$ & \textbf{59.37} & 6.83 & 5.38 & 0.01856 & 0.2767 & 0.8622 & 0.0108\\
 & MAFTC & 60.64 & 6.85 & 5.38 & 0.01874 & 0.2808 & 0.8657 & 0.0108\\
\addlinespace[2pt]
16 & Duo+DCD & 41.01 & 7.01 & 5.41 & 0.01910 & 0.2858 & 0.8535 & 0.0158\\
 & $\Psi$ & \textbf{40.17} & 7.00 & 5.41 & 0.01898 & 0.2840 & 0.8512 & 0.0166\\
 & MAFTC & \underline{40.67} & 7.03 & 5.40 & 0.01931 & 0.2879 & 0.8541 & 0.0171\\
\addlinespace[2pt]
32 & Duo+DCD & 34.07 & 7.12 & 5.40 & 0.01937 & 0.2914 & 0.8497 & 0.0216\\
 & $\Psi$ & \textbf{33.51} & 7.11 & 5.40 & 0.01930 & 0.2891 & 0.8493 & 0.0217\\
 & MAFTC & \underline{33.59} & 7.14 & 5.38 & 0.01971 & 0.2939 & 0.8515 & 0.0227\\
\addlinespace[2pt]
64 & Duo+DCD & 31.44 & 7.23 & 5.38 & 0.01968 & 0.2981 & 0.8500 & 0.0289\\
 & $\Psi$ & \underline{30.06} & 7.20 & 5.36 & 0.01956 & 0.2932 & 0.8459 & 0.0296\\
 & MAFTC & \textbf{29.93} & 7.23 & 5.32 & 0.01994 & 0.2972 & 0.8481 & 0.0329\\
\addlinespace[2pt]
128 & Duo+DCD & 29.77 & 7.32 & 5.32 & 0.01987 & 0.3027 & 0.8461 & 0.0407\\
 & $\Psi$ & \textbf{27.77} & 7.31 & 5.29 & 0.01980 & 0.2974 & 0.8390 & 0.0441\\
 & MAFTC & \underline{28.39} & 7.31 & 5.26 & 0.02005 & 0.3015 & 0.8437 & 0.0437\\
\bottomrule
\end{tabular}
\end{table}

\begin{table}[!ht]
\caption{\textbf{Complete text trajectory results.} Average cumulative token changes $L_{\rm act}$ and trajectory excess $R_{\rm traj}$ for the same 2,048 sequences per configuration as Table~\ref{tab:language-quality}. Net displacement is $D_{\rm net}=L_{\rm act}-R_{\rm traj}$.}
\label{tab:text-trajectories}
\centering
\small
\setlength{\tabcolsep}{6pt}
\begin{tabular}{c rr rr rr}
\toprule
 & \multicolumn{2}{c}{Duo+DCD} & \multicolumn{2}{c}{$\Psi$} & \multicolumn{2}{c}{MAFTC}\\
\cmidrule(lr){2-3}\cmidrule(lr){4-5}\cmidrule(lr){6-7}
NFE & $L_{\rm act}$ & $R_{\rm traj}$ & $L_{\rm act}$ & $R_{\rm traj}$ & $L_{\rm act}$ & $R_{\rm traj}$\\
\midrule
2 & 1261.5 & 238.3 & \underline{1260.6} & \textbf{237.3} & \textbf{1260.5} & \underline{237.5}\\
4 & \underline{2052.6} & \underline{1029.3} & 2056.5 & 1033.2 & \textbf{1178.2} & \textbf{198.3}\\
8 & \underline{2761.2} & \underline{1738.0} & 2782.8 & 1759.5 & \textbf{1114.3} & \textbf{142.8}\\
16 & \underline{3406.7} & \underline{2383.6} & 3465.6 & 2442.5 & \textbf{1067.2} & \textbf{96.9}\\
32 & \underline{4043.8} & \underline{3020.8} & 4177.3 & 3154.3 & \textbf{1042.2} & \textbf{71.5}\\
64 & \underline{4671.9} & \underline{3649.1} & 4955.5 & 3932.5 & \textbf{1031.8} & \textbf{59.4}\\
128 & \underline{5269.4} & \underline{4246.7} & 5849.2 & 4826.2 & \textbf{1028.8} & \textbf{54.2}\\
\bottomrule
\end{tabular}
\end{table}

\begin{figure}[!htbp]
\centering
\includegraphics[width=\linewidth]{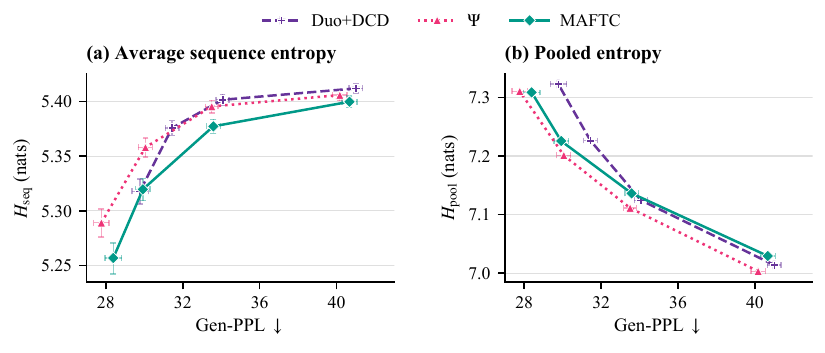}
\caption{\textbf{Two notions of generated-text entropy.}
Points show 16, 32, 64, and 128 NFE at fixed decoding settings, with seed 3407 and
2,048 sequences per configuration. Horizontal bars give 95\% confidence
intervals for Gen-PPL; vertical bars in (a) give intervals for average sequence entropy.
Pooled entropy in (b) is reported descriptively. Table~\ref{tab:paired-summary}
reports paired differences between methods.}
\label{fig:text-entropy}
\end{figure}

\FloatBarrier
\begin{samepage}
\paragraph{Complementary entropy measurements.}
Figure~\ref{fig:text-entropy} compares sequence and pooled unigram entropy
against Gen-PPL. Sequence entropy describes token diversity within each
output; pooling also incorporates differences in token usage across outputs.
For the $S$ equal-length outputs evaluated here, let $\widehat f_j$ be the
empirical token-frequency distribution of output $j$ and
$\bar f=S^{-1}\sum_{j=1}^{S}\widehat f_j$ the pooled distribution. Then
\[
H_{\rm pool}=H_{\rm seq}
+\frac1S\sum_{j=1}^{S}D_{\rm KL}(\widehat f_j\Vert\bar f),
\]
where $D_{\rm KL}$ is Kullback--Leibler divergence. The final term is the
equally weighted Jensen--Shannon divergence between the empirical
token-frequency distributions. Near Gen-PPL 30, it is 2.005, 1.843, and
1.906 nats for Duo+DCD at 128 NFE, $\Psi$ at 64 NFE, and MAFTC at 64 NFE,
respectively. Thus $\Psi$'s higher sequence entropy is outweighed by smaller
differences in token frequencies across outputs, yielding lower pooled
entropy. This complements the four-gram analysis below, which measures
coverage and repetition of ordered token combinations.
\end{samepage}

\begin{samepage}
\noindent\textbf{Four-gram decomposition.}
For $M$ equal-length sequences, let $u_s$ be the number of distinct four-grams in sequence $s$, $U$ the number of distinct four-grams across all sequences, and $T$ the total number of four-gram positions. Pooled Dist-4 is $D_4=U/T$, average sequence Rep-4 is $R_4=1-\sum_su_s/T$, and cross-sequence reuse is $C_4=(\sum_su_s-U)/T$. Hence $1-D_4=R_4+C_4$. For a comparison with either baseline, let $\Delta$ denote the MAFTC-minus-baseline difference. The change in corpus coverage then separates as
\[
\Delta D_4=-\Delta R_4-\Delta C_4.
\]
\end{samepage}
Figure~\ref{fig:fourgram-decomposition} compares these contributions at similar Gen-PPL, following Figure~\ref{fig:text-quality-diversity}. At three observed Gen-PPL levels near 41, 34, and 30, we pair each MAFTC configuration with the closest measured Gen-PPL of each baseline among NFE 16--128. All six pairs differ in Gen-PPL by less than 1.5\%; the figure reports both values and sampling budgets. Near Gen-PPL 41 and 34, MAFTC's higher pooled Dist-4 comes from less cross-sequence reuse despite slightly more within-sequence repetition. Near Gen-PPL 34, these differences coexist with about 75\% fewer token changes than either baseline (Table~\ref{tab:text-trajectories}).

\begin{figure}[!ht]
\centering
\includegraphics[width=\linewidth]{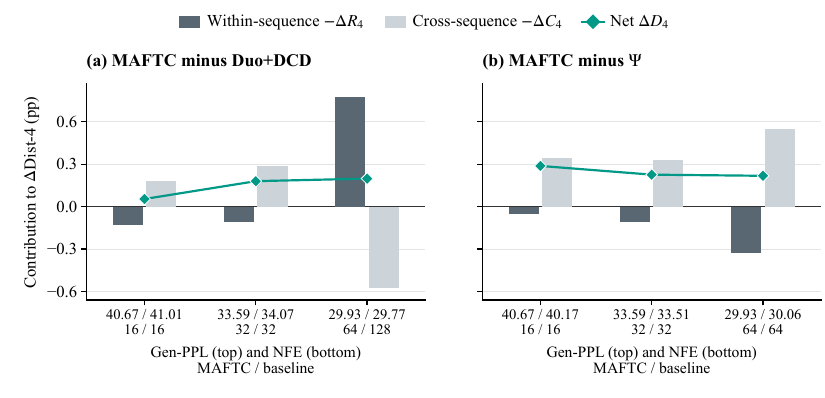}
\caption{\textbf{Sources of pooled Dist-4 differences at similar Gen-PPL.} Differences are MAFTC minus (a) Duo+DCD and (b) $\Psi$. Each horizontal label gives the measured Gen-PPL pair above its NFE pair, in MAFTC/baseline order. Bars show the contributions from within-sequence repetition ($-\Delta R_4$) and cross-sequence reuse ($-\Delta C_4$); diamonds show their sum, $\Delta D_4$. Values are in percentage points (pp). Positive contributions increase corpus four-gram coverage.}
\label{fig:fourgram-decomposition}
\end{figure}

Near Gen-PPL 30, the source of MAFTC's higher pooled Dist-4 depends on the baseline. Compared with Duo+DCD-128, MAFTC-64 has less within-sequence repetition but more cross-sequence reuse; compared with $\Psi$-64, the directions are reversed. For the MAFTC-64/Duo+DCD-128 pair, the differences $(\Delta D_4,\Delta R_4,\Delta C_4)$ in percentage points are $(+0.198,-0.773,+0.575)$ at seed 3407 and $(+0.314,-0.823,+0.510)$ at the independent seed 3408; the corresponding Gen-PPL differences are $+0.153$ and $+0.176$.

\FloatBarrier

\subsubsection{Paired comparisons for pretrained generators}
Table~\ref{tab:paired-summary} collects the main image and text comparisons.
Panel (a) uses the image budgets in Table~\ref{tab:generation-quality}; panel (b)
uses the same text budgets and adds 64 NFE, discussed in the quality--diversity
comparison. Both references are included at each displayed budget. Complete comparisons
against both baselines at all evaluated NFE budgets, including all entropy and
repetition intervals, are provided in \texttt{supplementary\_tables.zip}.

MAFTC has lower Gen-PPL than Duo+DCD at 64 and 128 NFE, but higher Gen-PPL
at 8 NFE and relative to $\Psi$ at 128 NFE. At 32 NFE, the Gen-PPL difference
from Duo+DCD is $-0.479$ with interval $[-0.950,-0.002]$, whose upper endpoint
is close to zero. Sequence entropy is lower at 32--128 NFE; repetition changes
depend on the budget and reference. These paired comparisons complement the
corpus-level coverage curves in Figure~\ref{fig:text-quality-diversity}.

\begin{table}[H]
\centering
\caption{\textbf{Key paired comparisons for pretrained generators.} Entries are
MAFTC minus the reference, with 95\% paired-bootstrap confidence intervals.
Image comparisons use 553 prompt clusters with four images each; text comparisons
use 2,048 complete sequences, all at seed 3407. GenEval and Rep-4 differences
are in percentage points; $H_{\rm seq}$ is in nats. The selected budgets are
specified above; full grids accompany the paper.}
\label{tab:paired-summary}
\fontsize{9}{10.8}\selectfont
\setlength{\tabcolsep}{4pt}
\begin{tabular}{rlrr}
\toprule
\multicolumn{4}{l}{\textbf{(a) Image quality and activity}}\\
NFE & Reference & $\Delta$ GenEval [95\% CI] & $\Delta$ NTC [95\% CI]\\
\midrule
8 & Euler & $+2.19\;[+0.34, +4.07]$ & $-266.8\;[-273.2, -260.3]$\\
8 & TC & $-0.99\;[-2.56, +0.63]$ & $-397.2\;[-403.5, -391.0]$\\
32 & Euler & $-0.88\;[-2.47, +0.72]$ & $-525.1\;[-535.2, -514.0]$\\
32 & TC & $-0.10\;[-1.90, +1.71]$ & $-567.8\;[-578.4, -556.6]$\\
128 & Euler & $+1.01\;[-0.69, +2.73]$ & $-551.0\;[-572.2, -527.0]$\\
128 & TC & $-0.23\;[-1.98, +1.52]$ & $-581.5\;[-602.6, -560.2]$\\
\bottomrule
\end{tabular}

\vspace{5pt}
\setlength{\tabcolsep}{3.5pt}
\begin{tabular}{rlrrrr}
\toprule
\multicolumn{6}{l}{\textbf{(b) Text quality, activity, and within-sequence diversity}}\\
NFE & Reference & $\Delta$ Gen-PPL & $\Delta$ NTC & $\Delta H_{\rm seq}$ & $\Delta$ Rep-4\\
\midrule
8 & Duo+DCD & \shortstack[r]{$+0.986$\\$[+0.338, +1.643]$} & \shortstack[r]{$-1646.9$\\$[-1648.5, -1645.3]$} & \shortstack[r]{$-0.0030$\\$[-0.0086, +0.0025]$} & \shortstack[r]{$+0.040$\\$[-0.017, +0.106]$}\\[3pt]
8 & $\Psi$ & \shortstack[r]{$+1.264$\\$[+0.637, +1.891]$} & \shortstack[r]{$-1668.4$\\$[-1670.1, -1666.8]$} & \shortstack[r]{$+0.0014$\\$[-0.0040, +0.0070]$} & \shortstack[r]{$-0.001$\\$[-0.055, +0.048]$}\\[3pt]
32 & Duo+DCD & \shortstack[r]{$-0.479$\\$[-0.950, -0.002]$} & \shortstack[r]{$-3001.5$\\$[-3003.8, -2999.3]$} & \shortstack[r]{$-0.0242$\\$[-0.0320, -0.0164]$} & \shortstack[r]{$+0.109$\\$[-0.008, +0.231]$}\\[3pt]
32 & $\Psi$ & \shortstack[r]{$+0.083$\\$[-0.384, +0.552]$} & \shortstack[r]{$-3135.1$\\$[-3137.4, -3132.7]$} & \shortstack[r]{$-0.0181$\\$[-0.0256, -0.0104]$} & \shortstack[r]{$+0.107$\\$[+0.009, +0.204]$}\\[3pt]
64 & Duo+DCD & \shortstack[r]{$-1.517$\\$[-2.049, -0.992]$} & \shortstack[r]{$-3640.1$\\$[-3642.8, -3637.5]$} & \shortstack[r]{$-0.0561$\\$[-0.0679, -0.0443]$} & \shortstack[r]{$+0.409$\\$[+0.206, +0.613]$}\\[3pt]
64 & $\Psi$ & \shortstack[r]{$-0.133$\\$[-0.639, +0.367]$} & \shortstack[r]{$-3923.7$\\$[-3926.5, -3920.8]$} & \shortstack[r]{$-0.0384$\\$[-0.0506, -0.0263]$} & \shortstack[r]{$+0.331$\\$[+0.139, +0.520]$}\\[3pt]
128 & Duo+DCD & \shortstack[r]{$-1.385$\\$[-1.986, -0.809]$} & \shortstack[r]{$-4240.7$\\$[-4243.8, -4237.6]$} & \shortstack[r]{$-0.0608$\\$[-0.0784, -0.0434]$} & \shortstack[r]{$+0.307$\\$[-0.011, +0.634]$}\\[3pt]
128 & $\Psi$ & \shortstack[r]{$+0.619$\\$[+0.021, +1.197]$} & \shortstack[r]{$-4820.4$\\$[-4823.9, -4816.9]$} & \shortstack[r]{$-0.0322$\\$[-0.0504, -0.0140]$} & \shortstack[r]{$-0.036$\\$[-0.360, +0.289]$}\\[3pt]
\bottomrule
\end{tabular}
\end{table}
\FloatBarrier

\subsubsection{Changing the coupling at fixed conditional marginals}
\label{app:coupling-ablations}

The activity decomposition suggests a direct intervention: change the coupling while preserving the prescribed conditional transport problem. At each interval and coordinate, we first draw a common endpoint $a$ using the same Duo prediction and then choose independently between the MAFTC and Duo+DCD ancestral kernels:
\begin{equation}
K_\lambda=(1-\lambda)K^{\rm MAFTC}+\lambda K^{\rm ancestral},
\qquad \lambda\in\{0,0.25,0.5,1\}.
\label{eq:conditional-coupling-interpolation}
\end{equation}
For a fixed endpoint $a$, both kernels map the same conditional input PMF $p$ to the same conditional output PMF $q$. Their mixture therefore also satisfies $pK_\lambda=q$. Since activity is linear in the coupling and MAFTC attains $\TV(p,q)$,
\begin{equation}
\chi(K_\lambda)=\lambda\chi(K^{\rm ancestral}).
\label{eq:conditional-excess-interpolation}
\end{equation}
This relation holds for each prescribed conditional pair; rollout activity and terminal quality also depend on the endpoint predictions encountered along the trajectory.

We keep the Duo-distilled checkpoint, time grid ($\delta=0.001$), top-$p=0.9$, and one prediction per interval fixed. Each cell contains 2,048 sequences of length 1,024 at seed 3407 with shared initial states. The $\lambda=0$ and $1$ controls reuse the original MAFTC and Duo+DCD benchmark outputs; the six interior cells are generated with the mixed kernel. Table~\ref{tab:coupling-quality} reports every cell and the key paired comparisons. Complete trajectory diagnostics and paired comparisons remain in \texttt{supplementary\_tables.zip}.

\begin{table}[H]
\centering
\fontsize{9}{10.8}\selectfont
\setlength{\tabcolsep}{4pt}
\renewcommand{\arraystretch}{1.05}
\caption{\textbf{Conditional coupling interpolation.} All cells use the same frozen model and conditional path; $\lambda=0$ is MAFTC and $\lambda=1$ is the Duo+DCD ancestral kernel. The upper panel combines generation statistics with NTC and average trajectory excess $R_{\rm traj}$. The lower panel gives $\lambda=0.25$ minus MAFTC: point estimates above 95\% paired-bootstrap CIs (Appendix~\ref{app:metric-definitions}).}
\label{tab:coupling-quality}
\label{tab:coupling-activity}
\label{tab:coupling-paired-ci}
\begin{tabular}{rrrrrrrr}
\toprule
NFE & $\lambda$ & Gen-PPL $\downarrow$ & $H_{\rm seq}$ & Dist-4 $\uparrow$ & Rep-4 $\downarrow$ & NTC & $R_{\rm traj}$\\
\midrule
32 & 0 & \textbf{33.591} & 5.3772 & 0.85151 & 0.02272 & \textbf{1042.25} & \textbf{71.49}\\
32 & 0.25 & \underline{33.594} & 5.3850 & 0.84928 & 0.02298 & \underline{1793.79} & \underline{801.12}\\
32 & 0.5 & 33.874 & 5.3889 & 0.84974 & 0.02237 & 2545.93 & 1538.80\\
32 & 1 & 34.070 & 5.4015 & 0.84971 & 0.02163 & 4043.76 & 3020.79\\
\midrule
64 & 0 & \textbf{29.926} & 5.3196 & 0.84810 & 0.03294 & \textbf{1031.82} & \textbf{59.38}\\
64 & 0.25 & \underline{30.525} & 5.3414 & 0.84912 & 0.03067 & \underline{1944.69} & \underline{948.49}\\
64 & 0.5 & 30.769 & 5.3525 & 0.84973 & 0.02964 & 2855.93 & 1845.63\\
64 & 1 & 31.443 & 5.3757 & 0.85001 & 0.02886 & 4671.91 & 3649.05\\
\midrule
128 & 0 & \textbf{28.388} & 5.2569 & 0.84369 & 0.04374 & \textbf{1028.76} & \textbf{54.25}\\
128 & 0.25 & 28.752 & 5.2773 & 0.84567 & 0.04210 & \underline{2093.05} & \underline{1093.14}\\
128 & 0.5 & \underline{28.688} & 5.2857 & 0.84325 & 0.04270 & 3155.89 & 2142.98\\
128 & 1 & 29.773 & 5.3178 & 0.84611 & 0.04067 & 5269.44 & 4246.73\\
\bottomrule
\end{tabular}

\medskip
\textbf{$\lambda=0.25$ minus MAFTC}\par\smallskip
\begin{tabular}{rrrr}
\toprule
NFE & $\Delta$Gen-PPL & $\Delta H_{\rm seq}$ & $\Delta$Rep-4\\
\midrule
32 & \shortstack[r]{$+0.003$\\$[-0.374,+0.378]$} & \shortstack[r]{$+0.0078$\\$[+0.0017,+0.0140]$} & \shortstack[r]{$+0.00026$\\$[-0.00058,+0.00110]$}\\
64 & \shortstack[r]{$+0.599$\\$[+0.185,+1.026]$} & \shortstack[r]{$+0.0217$\\$[+0.0127,+0.0309]$} & \shortstack[r]{$-0.00227$\\$[-0.00388,-0.00068]$}\\
128 & \shortstack[r]{$+0.365$\\$[-0.140,+0.879]$} & \shortstack[r]{$+0.0204$\\$[+0.0060,+0.0352]$} & \shortstack[r]{$-0.00164$\\$[-0.00435,+0.00102]$}\\
\bottomrule
\end{tabular}
\end{table}

Increasing $\lambda$ consistently increases observed NTC in these runs, whereas its terminal effects depend on the budget and statistic. At 64 NFE, $\lambda=0.25$ increases sequence entropy and decreases repetition, but also increases Gen-PPL. At 32 and 128 NFE, the confidence intervals for the Gen-PPL differences include zero (Table~\ref{tab:coupling-quality}, lower panel).

\FloatBarrier

\subsubsection{Decoding search and fixed-refresh controls}
\label{app:decoding-search}

On nonterminal intervals, the fixed-refresh control accepts an additional token proposal with a fixed probability $c$ when MAFTC retains the current token.

\paragraph{Search protocol.}
All evaluations use the same frozen Duo-distilled checkpoint, GPT-2-large scorer, 1,024-token length, batch size two, and generation seed 3407. We use separate samples for the pilot search and main evaluation: 315 pilot configurations each use 256 sequences, and 78 configurations in the main evaluation each use 2,048 sequences. Their random seeds are derived using offsets of 2,000,000 and zero, respectively. The pilot contains 162 $\Psi$, 117 fixed-refresh, and 36 MAFTC configurations across NFE 8, 16, and 32. Its base grid is top-$p\in\{0.8,0.9,1\}$ and temperature $T\in\{0.9,1,1.1\}$, with $\Psi$ strength $\eta\in\{0,0.05,0.1,0.2,0.4\}$ and fixed-refresh probability $c\in\{0.025,0.05,0.1\}$, followed by local midpoint refinements. Temperature divides raw Duo logits before softmax, nucleus truncation, and endpoint-posterior conversion.

For each NFE, pilot diversity targets are the observed Dist-4 values of MAFTC and the three fixed-refresh anchors at $p=0.9,T=1$. Within each family, we select the lowest-PPL pilot configuration meeting each target, deduplicate the selections, and include the anchors and default $\Psi$. Configurations and comparison pairs are frozen before the main evaluation. The main evaluation also includes a MAFTC scan over 19 decoding settings at each NFE: the nine-point base grid, five additional top-$p$ values $\{0.85,0.875,0.8875,0.925,0.95\}$ at $T=1$, and five temperatures $\{0.95,0.975,0.9875,1.025,1.05\}$ at $p=0.9$. The combined main evaluation therefore contains 57 MAFTC, nine $\Psi$, and 12 fixed-refresh configurations.

The accompanying \texttt{supplementary\_tables.zip} retains all 78 configurations from the main evaluation and 315 pilot configurations, including $H_{\rm seq}$, pooled unigram entropy, Dist-4, Rep-4, and NTC. Dist-4 values from the pilot and main evaluation are kept separate because pooled coverage depends on corpus size. Figure~\ref{fig:text-quality-diversity} compares coverage and repetition across NFE, while these search records show their dependence on decoding and refresh strength.

\paragraph{What additional refresh changes.}
The fixed-refresh control redistributes retained diagonal mass. For the same sampled endpoint, let $d_i=\min(p_i,q_i)$ and $D=\sum_i d_i$. On nonterminal intervals, draw a proposal $z\sim d/D$ independently of the baseline MAFTC transition; if MAFTC holds the current token $x$ and $z\ne x$, accept $z$ with probability $c$. This adds joint mass $c\,d_xd_z/D$ to each off-diagonal pair and removes the corresponding mass from the diagonal, preserving both prescribed conditional marginals. Writing $K_c$ for the resulting kernel, its conditional coupling excess is
\begin{equation}
\chi(K_c)=c\left(D-\frac{\sum_i d_i^2}{D}\right),\qquad D>0.
\label{eq:fixed-refresh-excess}
\end{equation}
If $D=0$, there is no retained diagonal mass to refresh. The terminal interval uses the original MAFTC update. The realized number of changes depends on the sampled proposals and trajectory. Table~\ref{tab:fixed-refresh-default} isolates this intervention at the paper's default decoding parameters, and Table~\ref{tab:psi-decoding-controls} includes all nine $\Psi$ settings from the main evaluation and the three other selected fixed-refresh settings, together with MAFTC controls using the same decoding settings.

At NFE 32, increasing fixed refresh from $c=0$ to $0.1$ lowers Gen-PPL from 33.591 to 28.376, but also lowers Dist-4 from 0.85151 to 0.83689 and raises Rep-4 from 0.02272 to 0.02565; NTC increases from 1042.25 to 3086.03. At $c=0.05$, the differences in Gen-PPL (refresh control minus MAFTC) at NFE 8, 16, and 32 are respectively $-0.828$, $-1.685$, and $-2.947$. Their 95\% paired-bootstrap CIs are $[-1.194,-0.455]$, $[-1.976,-1.395]$, and $[-3.251,-2.656]$; approximate Bonferroni-adjusted intervals across the 24 frozen comparisons are $[-1.416,-0.250]$, $[-2.160,-1.239]$, and $[-3.430,-2.482]$. The corresponding Dist-4 decreases at every NFE, and at NFE 32 the Rep-4 difference is $+0.00126$ with 95\% CI $[+0.00062,+0.00192]$.

At 32 NFE with $p=0.8,T=1.1$, tuned $\Psi$ ($\eta=0.2$) and fixed refresh ($c=0.1$) reach Gen-PPL 29.972 and 30.313 with Dist-4 0.85052 and 0.85128, respectively. They make about 5,674 and 3,109 token changes per sequence, respectively (Table~\ref{tab:psi-decoding-controls}).

\begin{table}[!htbp]
\centering
\fontsize{9}{10.8}\selectfont
\setlength{\tabcolsep}{5.5pt}
\renewcommand{\arraystretch}{1.05}
\caption{\textbf{Fixed refresh at identical default decoding parameters, $p=0.9,T=1$.} All nine configurations with $c>0$ and their MAFTC controls are shown. Increasing refresh lowers Gen-PPL and corpus coverage while increasing token changes.}
\label{tab:fixed-refresh-default}
\begin{tabular}{rlrrrr}
\toprule
NFE & Sampler & Gen-PPL $\downarrow$ & Dist-4 $\uparrow$ & Rep-4 $\downarrow$ & NTC\\
\midrule
8 & MAFTC & 60.639 & 0.86565 & 0.01081 & \textbf{1114.31}\\
8 & Refresh $c=0.025$ & 60.396 & 0.86413 & 0.01076 & \underline{1213.51}\\
8 & Refresh $c=0.05$ & \underline{59.811} & 0.86298 & 0.01072 & 1311.89\\
8 & Refresh $c=0.1$ & \textbf{59.427} & 0.86104 & 0.01112 & 1509.00\\
\midrule
16 & MAFTC & 40.667 & 0.85407 & 0.01710 & \textbf{1067.21}\\
16 & Refresh $c=0.025$ & 39.784 & 0.85304 & 0.01696 & \underline{1303.33}\\
16 & Refresh $c=0.05$ & \underline{38.982} & 0.85016 & 0.01702 & 1540.02\\
16 & Refresh $c=0.1$ & \textbf{37.515} & 0.84743 & 0.01760 & 2013.62\\
\midrule
32 & MAFTC & 33.591 & 0.85151 & 0.02272 & \textbf{1042.25}\\
32 & Refresh $c=0.025$ & 32.038 & 0.84724 & 0.02337 & \underline{1552.73}\\
32 & Refresh $c=0.05$ & \underline{30.645} & 0.84352 & 0.02399 & 2063.69\\
32 & Refresh $c=0.1$ & \textbf{28.376} & 0.83689 & 0.02565 & 3086.03\\
\bottomrule
\end{tabular}
\end{table}

\begin{table}[H]
\centering
\fontsize{9}{10.8}\selectfont
\setlength{\tabcolsep}{5.5pt}
\renewcommand{\arraystretch}{1.05}
\caption{\textbf{$\Psi$ and refresh controls under selected decoding settings.} Every $\Psi$ setting from the main evaluation is included. MAFTC values with default decoding settings appear in Table~\ref{tab:fixed-refresh-default}; the $p=0.8,T=1.1$ MAFTC anchors are shown here. The controls with the same decoding settings show differences in observed diversity across samplers.}
\label{tab:psi-decoding-controls}
\begin{tabular}{llrrrr}
\toprule
Sampler & Strength & Gen-PPL $\downarrow$ & Dist-4 $\uparrow$ & Rep-4 $\downarrow$ & NTC\\
\midrule
\multicolumn{6}{l}{NFE 8, $p=0.9,T=1$}\\
$\Psi$ & $\eta=0$ & \underline{59.522} & 0.86240 & 0.01082 & \textbf{2767.56}\\
$\Psi$ & $\eta=0.05$ & \textbf{59.375} & 0.86222 & 0.01082 & \underline{2782.76}\\
\midrule
\multicolumn{6}{l}{NFE 8, $p=0.8,T=1.1$}\\
MAFTC & --- & 68.279 & 0.88569 & 0.00814 & \textbf{1145.84}\\
Fixed refresh & $c=0.075$ & \underline{67.160} & 0.88390 & 0.00799 & \underline{1440.40}\\
$\Psi$ & $\eta=0.4$ & \textbf{60.899} & 0.87027 & 0.00900 & 3636.24\\
\midrule
\multicolumn{6}{l}{NFE 16, $p=0.9,T=1$}\\
$\Psi$ & $\eta=0.05$ & 40.168 & 0.85119 & 0.01655 & 3465.62\\
\midrule
\multicolumn{6}{l}{NFE 16, $p=0.8,T=1.1$}\\
MAFTC & --- & 44.461 & 0.87225 & 0.01367 & \textbf{1091.81}\\
Fixed refresh & $c=0.1$ & 40.862 & 0.86487 & 0.01441 & \underline{2038.71}\\
$\Psi$ & $\eta=0.3$ & \underline{36.738} & 0.85386 & 0.01488 & 4857.60\\
$\Psi$ & $\eta=0.4$ & \textbf{33.839} & 0.84528 & 0.01553 & 5807.03\\
\midrule
\multicolumn{6}{l}{NFE 32, $p=0.9,T=1$}\\
$\Psi$ & $\eta=0.05$ & 33.508 & 0.84925 & 0.02166 & 4177.32\\
\midrule
\multicolumn{6}{l}{NFE 32, $p=0.8,T=1.1$}\\
MAFTC & --- & 36.480 & 0.86687 & 0.02024 & \textbf{1064.95}\\
Fixed refresh & $c=0.1$ & \underline{30.313} & 0.85128 & 0.02263 & \underline{3108.75}\\
$\Psi$ & $\eta=0.15$ & 32.119 & 0.85585 & 0.02073 & 5037.87\\
$\Psi$ & $\eta=0.2$ & \textbf{29.972} & 0.85052 & 0.02166 & 5674.27\\
\bottomrule
\end{tabular}
\end{table}

\FloatBarrier

\FloatBarrier

\begingroup
\setlength{\intextsep}{6pt}
\subsection{Practical execution: query selection and measured cost}
\label{app:execution}
Selective execution uses the set of positions requiring an endpoint prediction to identify output rows for computation. Write $u=\alpha_t$, $v=\alpha_s$, and let $V$ denote vocabulary size. For the implemented uniform-path kernels, let $g_M$, $g_N$, and $g_\Psi$ denote the per-position query probabilities in one interval for MAFTC, Duo+DCD ancestral sampling, and $\Psi$, respectively. They are
\begin{align*}
g_M&=\frac{v-u}{1-u}, &
g_N&=g_M+\frac{1-v}{1-u}\frac{v-u}{v}\frac{(V-1)u}{1+(V-1)u},\\
g_\Psi&=\omega_k g_N+(1-\omega_k)v.
\end{align*}
Here $\omega_k$ is the baseline's existing mixture weight, with $\eta=0.05$ and terminal $\omega_k=1$. For $0\le u<v\le1$, $g_M\le g_N,g_\Psi$: the additional term is nonnegative and $v-g_M=u(1-v)/(1-u)\ge0$. Using the same uniform random variables for the gates makes MAFTC's queried positions a subset of those queried by either baseline; applying the same nondecreasing padding rule preserves this ordering of row counts. A queried position need not change, while the baselines can change tokens without an endpoint prediction: query counts and NTC are distinct.

We compare dense output computation (\emph{full}), selection at the vocabulary head (\emph{head}), and selection at the final block plus head (\emph{tail}). Selected attention queries retain all keys, values, and masks from the current full sequence, before any update that does not require an endpoint prediction. In exact arithmetic, these operations produce the same outputs at the selected positions as dense execution. The first 11 blocks and the final block's fused QKV projection remain dense. Each variant's own outputs are evaluated because BF16 rounding can depend on matrix shape.

\paragraph{Shared implementation and accounting.}
All arms use the released 169.6M-parameter Duo-distilled checkpoint, seed 3407, batch size one, length 1,024, top-$p=0.9$, SDPA, compilation, and CUDA Graphs, with stored FP32 weights, internal BF16, and TF32 disabled. The \emph{full} variant uses the same exact query gate. Within each sampler, all three variants apply probability processing and top-$p$ filtering only to the same padded selected rows. Table~\ref{tab:execution_queries} sums selected token positions over all stages of each request (Query rows); Padded rows also includes rows added for matrix execution. Both totals are averaged over the 512 trajectories used for quality evaluation. Our FLOP count includes major linear and attention products, including padding, with a multiply-add counted as two operations. Normalization, nonlinearities, sorting, RNG, and transfers are included in latency but excluded from this matrix count. Empty query sets skip a model call.

\begin{table}[htbp]\centering
\fontsize{9}{10.2}\selectfont
\setlength{\tabcolsep}{3.5pt}\renewcommand{\arraystretch}{1.04}
\caption{Query accounting at 64/128 NFE for tail execution, averaged over its 512 trajectories used for quality evaluation. These FLOP counts correspond to the same trajectories; timing requests are separate. Calls equal scheduled NFE except two MAFTC trajectories at NFE 128, each using 127 calls (average 127.996).}
\label{tab:execution_queries}
\begin{tabular}{rlrrrr}
\toprule
NFE & Sampler & NTC & Query rows & Padded rows & TFLOPs \\
\midrule

64 & Duo+DCD & \underline{4668.4} & \underline{10338.0} & \underline{11961.7} & \underline{13.793} \\

64 & $\Psi$ & 4894.3 & 10693.1 & 12487.1 & 13.841 \\

64 & MAFTC & \textbf{1031.4} & \textbf{5767.6} & \textbf{6884.8} & \textbf{13.332} \\

\midrule

128 & Duo+DCD & \underline{5266.3} & \underline{11635.5} & \underline{14295.9} & \underline{26.711} \\

128 & $\Psi$ & 5733.0 & 12459.4 & 15367.5 & 26.809 \\

128 & MAFTC & \textbf{1030.0} & \textbf{6381.8} & \textbf{8864.9} & \textbf{26.216} \\

\bottomrule
\end{tabular}
\end{table}

The dense blocks dominate the matrix count, while latency also includes query-dependent probability processing and sampling.

\paragraph{Quality and timing protocol.}
Each of the 36 execution configurations has 512 outputs, evaluated using the metrics and token handling in Appendix~\ref{app:metric-definitions}. This dedicated rerun uses seed 3407 with a separate per-request random-stream construction from the main 2,048-output evaluation. Table~\ref{tab:execution-ablation} reports all full/head/tail variants at the 64/128 NFE budgets used in the main cost comparison. The complete 16/32/64/128 NFE results, including pooled entropy and throughput, are provided in \texttt{supplementary\_tables.zip}.

On an otherwise unused RTX 4090, each configuration uses 16 independent timing IDs with five repetitions of each complete generation request; variant order rotates. We report latency in milliseconds and TFLOPs from matrix operations per request, both averaged over these 80 timed requests. Latency includes query selection, model evaluation, probability processing, sampling, numerical validation, activity bookkeeping, and output transfer. Loading, compilation, and warm-up are excluded. Throughput is the total number of final token positions divided by total elapsed time.

We use the paired-bootstrap procedure in Appendix~\ref{app:metric-definitions}. Quality intervals resample complete sample IDs and recompute token-weighted Gen-PPL; timing intervals resample the 16 independent ID groups, retaining all five repetitions.

\begin{table}[H]\centering
\fontsize{9}{10.2}\selectfont
\setlength{\tabcolsep}{2.4pt}\renewcommand{\arraystretch}{1.04}
\caption{Execution ablation at 64/128 NFE. (a) Rankings compare samplers at the same NFE and execution variant. Quality, diversity, and NTC use each variant's own 512 outputs; latency and TFLOPs from matrix operations are averaged over the 80 timed requests. All predictors retain full sequence context. (b) MAFTC minus each reference under common tail execution, with 95\% paired-bootstrap CIs.}
\label{tab:execution-ablation}
\label{tab:execution_quality}
\label{tab:execution_cost}
\label{tab:execution_differences}
\begin{tabular}{rlrrrrrrr}
\toprule
\multicolumn{9}{l}{\textbf{(a) Quality, activity, and cost}}\\
NFE & Sampler/execution & Gen-PPL & $H_{\rm seq}$ & Dist-4 & Rep-4 & NTC & ms & TFLOPs \\
\midrule
64 & Duo+DCD/full & \underline{30.93} & 5.3700 & 0.8981 & 0.0270 & \underline{4670.0} & \underline{346.44} & \textbf{18.667} \\
64 & Duo+DCD/head & \underline{30.97} & 5.3699 & 0.8983 & 0.0268 & \underline{4670.2} & \underline{318.26} & \underline{14.528} \\
64 & Duo+DCD/tail & 31.41 & 5.3826 & 0.8970 & 0.0281 & \underline{4668.4} & \underline{314.02} & \underline{13.790} \\
\addlinespace[2pt]
64 & $\Psi$/full & \textbf{30.42} & 5.3765 & 0.8967 & 0.0271 & 4895.5 & 350.62 & \textbf{18.667} \\
64 & $\Psi$/head & \textbf{30.37} & 5.3745 & 0.8967 & 0.0272 & 4895.9 & 322.72 & 14.570 \\
64 & $\Psi$/tail & \underline{31.05} & 5.3772 & 0.8990 & 0.0273 & 4894.3 & 318.53 & 13.840 \\
\addlinespace[2pt]
64 & MAFTC/full & 31.03 & 5.3349 & 0.8997 & 0.0319 & \textbf{1031.8} & \textbf{307.44} & \textbf{18.667} \\
64 & MAFTC/head & 31.04 & 5.3367 & 0.8998 & 0.0318 & \textbf{1031.9} & \textbf{278.17} & \textbf{14.141} \\
64 & MAFTC/tail & \textbf{30.59} & 5.3353 & 0.8978 & 0.0318 & \textbf{1031.4} & \textbf{273.16} & \textbf{13.335} \\
\midrule
128 & Duo+DCD/full & 29.74 & 5.3318 & 0.8928 & 0.0374 & \underline{5267.8} & \underline{623.68} & \textbf{37.333} \\
128 & Duo+DCD/head & 29.79 & 5.3316 & 0.8929 & 0.0373 & \underline{5267.8} & \underline{564.52} & \underline{28.320} \\
128 & Duo+DCD/tail & 29.53 & 5.3214 & 0.8927 & 0.0366 & \underline{5266.3} & \underline{554.36} & \underline{26.713} \\
\addlinespace[2pt]
128 & $\Psi$/full & \textbf{28.78} & 5.3156 & 0.8937 & 0.0373 & 5733.0 & 627.46 & \textbf{37.333} \\
128 & $\Psi$/head & \textbf{28.79} & 5.3165 & 0.8934 & 0.0373 & 5733.0 & 568.21 & 28.393 \\
128 & $\Psi$/tail & \textbf{27.84} & 5.3110 & 0.8919 & 0.0379 & 5733.0 & 558.34 & 26.800 \\
\addlinespace[2pt]
128 & MAFTC/full & \underline{29.47} & 5.2776 & 0.8959 & 0.0420 & \textbf{1028.1} & \textbf{576.46} & \textbf{37.333} \\
128 & MAFTC/head & \underline{29.45} & 5.2790 & 0.8959 & 0.0417 & \textbf{1028.1} & \textbf{516.61} & \textbf{27.898} \\
128 & MAFTC/tail & \underline{28.87} & 5.2555 & 0.8923 & 0.0457 & \textbf{1030.0} & \textbf{505.98} & \textbf{26.216} \\
\bottomrule
\end{tabular}

\vspace{5pt}
\begin{tabular}{rlrr}
\toprule
\multicolumn{4}{l}{\textbf{(b) Paired tail-execution comparisons}}\\
NFE & Reference & $\Delta$ Gen-PPL [95\% CI] & $\Delta$ latency, ms [95\% CI] \\
\midrule
64 & Duo+DCD & $-0.81\;[-1.87, +0.23]$ & $-40.87\;[-42.07, -39.66]$ \\
64 & $\Psi$ & $-0.46\;[-1.52, +0.60]$ & $-45.37\;[-46.76, -43.90]$ \\
\midrule
128 & Duo+DCD & $-0.67\;[-1.82, +0.49]$ & $-48.38\;[-48.94, -47.79]$ \\
128 & $\Psi$ & $+1.02\;[-0.10, +2.18]$ & $-52.36\;[-53.10, -51.62]$ \\
\bottomrule
\end{tabular}
\end{table}

At 64/128 NFE, MAFTC reduces latency and matrix work under common tail execution. MAFTC has lower Gen-PPL than both baselines at 64 NFE and lies between Duo+DCD and $\Psi$ at 128 NFE; the intervals for all four Gen-PPL differences include zero.

\paragraph{Additional scheduling controls.}
For the 16/32 NFE controls, we precompute the queried positions across steps at the start of each request and capture model evaluation, sampling, state updates, and activity recording for each step in a CUDA Graph. These optimizations reproduce all 512 reference outputs, every stage's activity and query counts, and all states in eight saved trajectories per sampler/budget, with unchanged FLOP counts. The supplementary archive contains all 18 scheduling configurations and 12 paired timing comparisons, measured in one session using 16 IDs with five repetitions.
\endgroup

\FloatBarrier

\subsection{Exact-posterior evaluation of complete sampling distributions}\label{app:oracle-accuracy}
\paragraph{Targets and conditional paths.}
We enumerate $\mathcal X=\{0,1,2,3\}^3$, containing 64 sequence states. For each seed in $\{0,\ldots,7\}$, a NumPy random generator draws a token distribution $b\sim\operatorname{Dirichlet}(2,2,2,2)$, a row-stochastic matrix $R$ with independent $\operatorname{Dirichlet}(1,1,1,1)$ rows, and a uniformly random permutation matrix $P$. We vary the dependence coefficient $\zeta\in\{0,0.5,0.9\}$ and define the target by
\[
M_\zeta=(1-\zeta)\mathbf1 b^\top+\zeta(0.15R+0.85P),\qquad
p_1(a)=b(a^1)M_\zeta(a^2\mid a^1)M_\zeta(a^3\mid a^2).
\]
Rows of $M_\zeta$ index the preceding token. At $\zeta=0$, the target factorizes as $b^{\otimes3}$; the other settings introduce coordinate dependence. All samplers share the same target and geometry at each seed.

Both paths factorize conditional on the endpoint. For one coordinate, the uniform-source path is $p_t(x\mid a)=t\1\{x=a\}+(1-t)/4$. The Gibbs path is $p_t(x\mid a)\propto\exp[-\beta(t)\rho(x,a)]$, with $\beta(t)=3t/(1-t)$. To construct its geometry, we draw four five-dimensional standard Gaussian embeddings from the same generator and normalize them to obtain unit vectors $e_0,\ldots,e_3$. We set the diagonal distances to zero and divide $(2-2\langle e_x,e_a\rangle)^2$ by its average over off-diagonal pairs. The controlled experiment uses this differentiable schedule at $t=0$, whereas pretrained FUDOKI uses the schedule in Appendix~\ref{app:conditional-evaluation}. Both paths begin at the uniform source and end at the sampled data endpoint.

\paragraph{Exact posterior and parallel updates.}
For each current sequence $x$, we compute the full endpoint posterior by Bayes' rule and then obtain its token marginals by summation. The implemented parallel kernel is
\[
K_{t,s}(y\mid x)=\prod_{i=1}^3\sum_{a^i\in\mathcal T}
p_{1\mid t}^{\,i}(a^i\mid x)K^{a^i}_{t,s}(y^i\mid x^i),
\]
where $K^{a^i}_{t,s}$ is the sampler's conditional coordinate kernel. Thus the experiment removes posterior prediction error while retaining finite-step and parallel-update effects. The parallel kernel marginalizes independent coordinatewise endpoint draws from the exact token posteriors. In the final interval, each coordinate is sampled from its current token posterior.

We use uniform time grids $t_k=k/N$, with $N\in\{1,2,4,8,16,32,64,128\}$. On the uniform path, we compare the conditional ancestral bridge, the $\Psi$ mixture with rescale parameter $\eta=0.05$, and MAFTC. On the Gibbs path, we compare the first-event exponential Euler update, the time-corrected update, and MAFTC. The baseline constructions follow Appendix~\ref{app:baseline-definitions}, with $\alpha_k=t_k$ for the uniform path. For a nonterminal first interval at $t_0=0$, we set $\sigma_0=\eta$; the ancestral and resampling kernels coincide there, so this choice does not alter the first update. For this Gibbs schedule, $\dot\beta(0)=3$. TC approximates the first-jump distribution using four quadrature nodes while keeping the source state fixed. All methods use one posterior query per grid interval, with no truncation or temperature adjustment. NFE counts posterior queries along a rollout.

\paragraph{Measurements and numerical checks.}
Starting from the uniform source, we propagate the complete sequence distribution in double precision and report $\TV(\widehat p_1,p_1)$, where $\widehat p_1$ is the terminal sampler law. Expected cumulative Hamming activity is summed under the evolving sampler law, and net displacement is computed by propagating the joint law of initial and current states. We verify stochastic kernel rows, nonnegative entries, mass conservation, and agreement between the propagated marginal and the initial/current joint law. The sweep contains 1,152 configurations: eight seeds, three dependence settings, two paths, eight NFEs, and three methods per path. Lines show averages over the eight target/geometry seeds, and bands show interquartile ranges.

\begin{figure}[p]
\centering
\includegraphics[width=\linewidth]{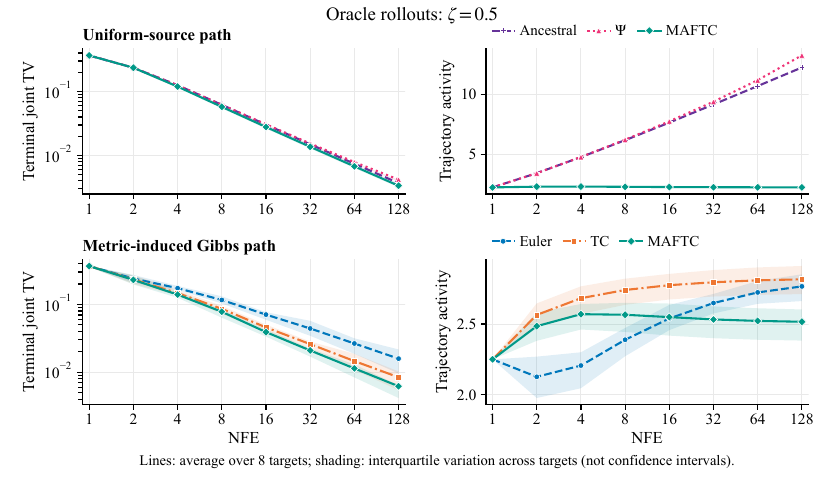}
\caption{\textbf{Exact-posterior sampling with moderate dependence ($\zeta=0.5$).} Complete NFE curves for both uniform and Gibbs paths, including NFE 1 and 2; the Gibbs accuracy--activity relation also appears in Figure~\ref{fig:real-models}(b). Lines and bands show averages and interquartile ranges over seeds; terminal joint TV uses a logarithmic scale.}
\label{fig:oracle-moderate}
\end{figure}

Figure~\ref{fig:oracle-moderate} reports terminal accuracy and activity across the full NFE grid at moderate dependence. On the uniform path, MAFTC reaches similar terminal TV with much lower activity: at 128 NFE, the activity is 2.25, versus 12.23 for ancestral sampling. On the Gibbs path, coarse Euler steps can have lower activity but greater terminal error; at 32--128 NFE, MAFTC lowers both relative to Euler and TC.

\begin{figure}[p]
\centering
\includegraphics[width=\linewidth]{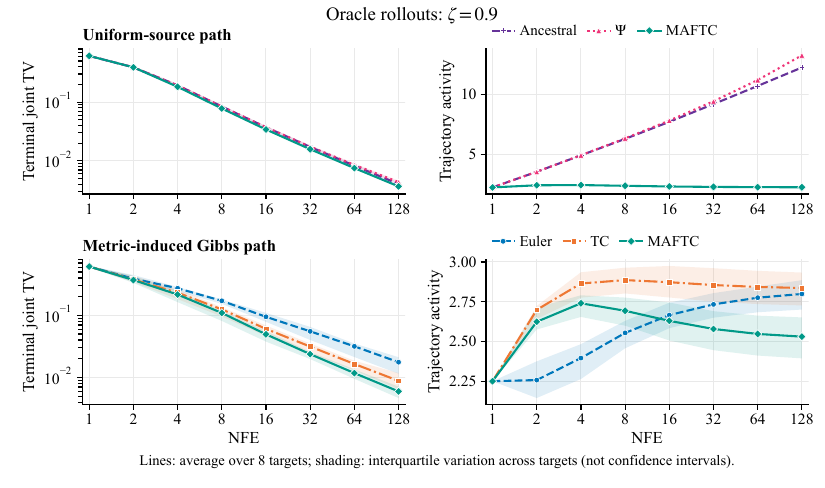}
\caption{\textbf{Exact-posterior sampling with strong dependence ($\zeta=0.9$).} Lines and bands show averages and interquartile ranges over eight targets. The comparison retains the same path-specific methods and grid as the moderate-dependence setting.}
\label{fig:oracle-strong}
\end{figure}

\paragraph{Dependence and accumulated error.}
Figures~\ref{fig:oracle-moderate} and~\ref{fig:oracle-strong} report the full NFE curves for dependent targets. For independent targets, uniform-path ancestral sampling, $\Psi$, and MAFTC, as well as Gibbs-path MAFTC, attain terminal TV below $3\times10^{-14}$ across the sweep; Gibbs-path Euler and TC retain finite-step error. With dependent targets, exact token posteriors no longer make the parallel finite-step update exact, and terminal TV decreases under grid refinement.

Using the prescribed full-sequence marginals, we define the summed local TV as
\[
\sum_{k=0}^{N-1}\TV\!\left(p_{t_k}K_{t_k,t_{k+1}},p_{t_{k+1}}\right).
\]
This quantity and terminal TV can rank samplers differently. At seed 0, $\zeta=0.9$, and two Gibbs steps, MAFTC has a larger summed local TV than TC (0.5526 versus 0.5485), yet a smaller terminal TV (0.3541 versus 0.3646). This reversal in sampler rankings shows that terminal accuracy must be assessed from the propagated output distribution.

\end{document}